\pdfoutput=1

\documentclass[10pt,journal,compsoc]{IEEEtran}
\ifCLASSOPTIONcompsoc
  \usepackage[nocompress]{cite}
\else
  \usepackage{cite}
\fi
\ifCLASSINFOpdf
\else
\fi
\ifCLASSOPTIONcompsoc
 \usepackage[caption=false,font=footnotesize,labelfont=sf,textfont=sf]{subfig}
\else
 \usepackage[caption=false,font=footnotesize]{subfig}
\fi
\usepackage[utf8]{inputenc} % allow utf-8 input
\usepackage[T1]{fontenc}    % use 8-bit T1 fonts
\usepackage{hyperref}       % hyperlinks
\usepackage{url}            % simple URL typesetting
\usepackage{booktabs}       % professional-quality tables
\usepackage{amsfonts}       % blackboard math symbols
\usepackage{nicefrac}       % compact symbols for 1/2, etc.
\usepackage{microtype}      % microtypography
\usepackage{multirow}
\usepackage[inline]{enumitem}
\usepackage{amsmath}
\usepackage{amssymb}
\usepackage{xcolor}
\usepackage{subfig}
\usepackage{graphicx}
\usepackage{tipa}
\usepackage[linesnumbered,ruled,vlined]{algorithm2e}
\usepackage{amsfonts}
\newcommand{\hjk}[1]{\textcolor{black}{#1}}
\newcommand{\sjy}[1]{\textcolor{black}{#1}}
\newcommand{\ysj}[1]{\textcolor{black}{#1}}
\newcommand{\concat}[2]{\underset{#1=1}{\overset{#2}{\text{\Large{\textdoublevertline}}}}}
\newcommand{\icml}[1]{\textcolor{black}{#1}}
\newcommand{\invhatD}[1]{\left (\hat{D}^{(#1)} \right )^{-1}}
\newcommand{\tpami}[1]{\textcolor{black}{#1}}

\newtheorem{proposition}{Proposition}
\def\Rb{\textbf{R}}
\newcommand{\alphab}{\mathbb{\alpha}}
\newcommand{\invD}[1]{\left (D^{(#1)} \right )^{-1}}
\begin{document}
\title{Graph Transformer Networks: Learning Meta-path Graphs to Improve GNNs}

\author{Seongjun~Yun, Minbyul~Jeong, Sungdong Yoo, Seunghun Lee, Sean S. Yi, , Raehyun~Kim \\ Jaewoo~Kang, Hyunwoo~J.~Kim
\IEEEcompsocitemizethanks{\IEEEcompsocthanksitem S. Yun, M. Jeong, S. Yoo, S. Lee, S. Yi, R. Kim, J. Kang and H. J. Kim are with the Department of Computer Science and Engineering, Korea University, Seoul 02841, Korea.\protect\\
E-mail: \{ysj5419, minbyuljeong, ysd424, llsshh319, seanswyi, raehyun, kangj and hyunwoojkim\}@korea.ac.kr}
\IEEEcompsocitemizethanks{\IEEEcompsocthanksitem Hyunwoo. J. Kim and Jaewoo Kang are the corresponding authors.}
}

\IEEEtitleabstractindextext{%
\begin{abstract}
\tpami{Graph Neural Networks (GNNs) have been widely applied to various fields due to their powerful representations of graph-structured data.}
Despite the success of GNNs, most existing GNNs are designed to learn node representations on the \textit{fixed} and \textit{homogeneous} graphs. 
The limitations especially become problematic when learning representations on a misspecified graph or a \emph{heterogeneous} graph that consists of various types of nodes and edges. 
To address this limitations, we propose Graph Transformer Networks (GTNs) that are capable of generating new graph structures, \tpami{which preclude noisy connections and include useful connections (e.g., meta-paths) for tasks, while learning effective node representations on the new graphs in an end-to-end fashion. We further propose enhanced version of GTNs, Fast Graph Transformer Networks (FastGTNs), that improve scalability of graph transformations.
Compared to GTNs, FastGTNs are 230$\times$ faster and use 100$\times$ less memory while allowing the \textit{identical} graph transformations as GTNs.}
In addition, we extend graph transformations to the semantic proximity of nodes allowing \textit{non-local} operations beyond meta-paths.
Extensive experiments on both homogeneous graphs and heterogeneous graphs show that GTNs and FastGTNs with non-local operations achieve the state-of-the-art performance for node classification tasks. 

The code is available: \url{https://github.com/seongjunyun/Graph_Transformer_Networks}
\end{abstract}
}

\maketitle

% is not selected <OR> if conference mode is selected - because compsoc
% conference papers position the abstract like regular (non-compsoc)
\IEEEdisplaynontitleabstractindextext

% \fi
\IEEEpeerreviewmaketitle

\IEEEraisesectionheading{\section{Introduction}\label{sec:introduction}}
Graph Neural Networks (GNNs) have become an increasingly popular tool to learn the representations of graph-structured data. 
They are widely used in a variety of tasks such as node classification \cite{kipf2016gcn, li2018deeper, xu2018jknet, liu2019geniepath, liu2020deepergnn}, link prediction \cite{kipf2016variational, schlichtkrull2018rgcn, zhang2018link, teru2019inductive, vashishth2019compgcn}, graph classification \cite{ying2018diffpool, xu2018powerful, gao2019graphunet, ma2019graph}, and graph generation \cite{you2018graphrnn, li2018generative, you2018graph, liao2019efficient, dai2020scalable}. 

Despite their effectiveness to learn representations on graphs, most GNNs assume that the given graphs are fixed and homogeneous. 
% This is problematic especially when the graph structure is noisy.
Since the graph convolutions discussed above are determined by a fixed graph structure, a noisy graph with missing/spurious connections results in ineffective convolution with wrong neighbors on the graph.
In addition, in some applications constructing a graph to operate GNNs is not trivial.
For example, a citation network has multiple types of nodes (e.g., authors, papers, conferences) and edges defined by their relations (e.g., author-paper, paper-conference), which referred to as \emph{heterogeneous} graphs. In heterogeneous graphs, the importance of each node type and edge type can vary depending on the task, and some node/edge types may even become completely useless.
A na\"ive approach to deal with the heterogeneous graphs is to ignore the node/edge types and treat them as in a \emph{homogeneous} graph (a standard graph with one type of nodes and edges).
This, apparently, is suboptimal since models cannot exploit the type information.
A more recent remedy is to manually design useful meta-paths, which are paths connected with heterogeneous edge types, and transform a heterogeneous graph into a \emph{homogeneous} graph defined by the meta-paths. Then conventional GNNs can operate on the transformed homogeneous graphs  \cite{wang2019han, zhang2018deep}.
% \new{while also being able to take advantage of the rich information of heterogeneous graphs}
Despite improving on previous approaches, this is a two-stage approach and requires hand-crafted meta-paths for each problem. The accuracy of downstream analysis can be significantly affected by the choice of these meta-paths.

\tpami{To address this limitation, we develop the Graph Transformer Networks (GTNs) that learn to transform an original graph into new graphs that preclude noisy connections and include useful multi-hop connections (e.g., meta-paths) for each task, and learn node representation on the new graphs in an end-to-end fashion. Specifically, the Graph Transformer layer, a core layer of GTN, learns a soft selection of adjacency matrices for edge types and multiply two selected adjacency matrices to generate useful meta-paths.} \hjk{Also, by leveraging an identity matrix, GTN can generate new graph structures based on arbitrary-length composite relations connected with softly chosen edge types in a heterogeneous graph.}

\hjk{Furthermore, we address the scalability issue of GTNs. To transform graphs, GTNs \textit{explicitly} compute a new adjacency matrix of meta-paths by the matrix multiplications of huge adjacency matrices. }
\tpami{This requires substantial computational costs and large memory making it infeasible to apply GTNs to a large graph. 
To address this issue, we propose an enhanced version of GTNs, Fast Graph Transformer Networks (FastGTNs), that \textit{implicitly} transform the graphs without the multiplication of two adjacency matrices. 
Compared to GTNs, FastGTNs are 230$\times$ faster and use 100$\times$ less memory while allowing the \textit{identical} graph transformations as GTNs.}

\hjk{Another issue of GTNs is its edge generation is limited to the nodes connected by a meta-path of the input graphs, which do not take into account the semantic proximity of nodes. We further extend graph transformations to the semantic proximity of nodes allowing \textit{non-local} operations beyond meta-paths.}

\tpami{The new graph structures from GTNs and FastGTNs lead to effective node representations resulting in state-of-the-art performance, without any predefined meta-paths from domain knowledge, on six benchmark classification on heterogeneous graphs.} \hjk{In addition, since GTNs and FastGTNs learn variable lengths of useful meta-paths (i.e., the neighborhood range of each node), GTNs and FastGTNs achieve state-of-the-art performance on six homogeneous graph datasets by adjusting neighborhood ranges for each dataset.}

Our \textbf{contributions} are as follows:
\begin{enumerate}[label=(\roman*)]
    \item We propose a novel framework Graph Transformer Networks, to learn a new graph structure which involves identifying useful meta-paths and multi-hop connections for learning effective node representation on graphs.
    \item \tpami{We propose Fast Graph Transformer Networks (FastGTNs) that \textit{implicitly} transform the graphs without the multiplication of adjacency matrices which requires excessive resources while allowing the identical transformations of GTNs.}
    \item \tpami{We extend graph transformations to non-local operations incorporating the node features to utilize the semantic proximity of nodes beyond meta-paths.}
    \item \tpami{We prove the effectiveness of node representation learnt by Graph Transformer Networks and FastGTNs with non-local operations resulting in the best performance against state-of-the-art GNN-baselines in six benchmark node classification on heterogeneous graphs and six benchmark node classification on homogeneous graphs. Also, our experiments demonstrate that on a large graph dataset FastGTNs show 230x faster inference time and 100x less memory usage than the GTNs.}

\end{enumerate}

\section{Related Works}
\tpami{Recent years have witnessed significant development in deep learning architectures for graphs.
\cite{bruna2013spectral} first proposed a convolution operation on graphs leveraging the Fourier transform and convolution kernels in a spectral domain.
\cite{defferrard2016convolutional,li2018deeper} extended and improved spectral-based Graph Convolutional Networks (GCNs).
On the other hand, spatial-based GNNs \cite{hamilton2017inductive, velivckovic2017gat, xu2018powerful, xu2018jknet, abu2019mixhop} have been proposed to improve the scalability of GNNs by 
performing graph convolution operations in the graph domain. 
\cite{kipf2016gcn} proposed the first-order approximation of the spectral filter using the Chebyshev polynomials.
GCNs, the simplified spectral-based GNNs, can be viewed as spatial-based GNNs as well.
Graph Attention Networks (GATs) \cite{velivckovic2017gat} incorporate the attention mechanism into GNNs, which differentially aggregate the representations of neighbors on graphs using attention scores.
GraphSAGE \cite{hamilton2017inductive} expanded the operating range of GNNs for the inductive setting generating representations for unseen nodes by variable aggregation operations.
Jumping Knowledge Networks (JKNets) \cite{xu2018jknet} utilized flexible neighborhood ranges by adopting the skip-connection and Mixhop \cite{abu2019mixhop} leveraged a combination of powers of normalized adjacency matrices to aggregate features at various distances.}

\tpami{In contrast to the above works on \textit{homogeneous} graphs, 
several studies \cite{schlichtkrull2018rgcn, hu2020hgt, wang2019han} have attempted to extend GNN architectures to \textit{heterogeneous} graphs that contain multiple types of nodes and edges.
They are categorized into two approaches: GNNs with relation-specific parameters \cite{schlichtkrull2018rgcn, hu2020hgt} and GNNs with relation-based graph transformations \cite{wang2019han}. 
Relational Graph Convolutional Networks (R-GCNs) \cite{schlichtkrull2018rgcn} employed GCNs with relation-specific convolutions (or weight matrices) to deal with heterogeneous graphs. 
\cite{hu2020hgt} proposed the Heterogeneous Graph Transformer (HGT) to parameterize the meta relation triplet of each edge type and used a structure that utilizes the self-attention of the transformer architecture \cite{vaswani2017attention} to learn specific patterns of different relationships.
The second approaches, GNNs with relation-based graph transformations, generally utilize meta-paths. 
The Heterogeneous Graph Attention Network (HAN)  \cite{wang2019han} first transforms heterogeneous graphs into homogeneous graphs using manually selected meta-paths and applies an attention-based GNN on the graphs. 
However, the HAN has limitations that it is a multi-stage approach and requires the manual selection of meta-paths in each dataset. Also, performance can be significantly affected by the choice of meta-paths. Unlike this approach, our Graph Transformer Networks can operate on a heterogeneous graph and transform the graph for tasks while learning node representations on the transformed graphs in an end-to-end fashion. }

\section{Method}
The goal of our framework, Graph Transformer Networks, is to generate new graph structures and learn node representations on the learned graphs simultaneously. Unlike most GNNs on graphs that assume the graph is given, GTNs seek for new graph structures
using multiple candidate adjacency matrices to perform more effective graph convolutions and learn more powerful node representations. 
Learning new graph structures involves identifying useful meta-paths, which are paths connected with heterogeneous edges, and multi-hop connections. 
Before introducing our framework, we first briefly review the basic notions of heterogeneous graphs and GCN, and then introduce our Graph Transformer Networks.

%\subsection{Preliminaries: Heterogeneous Graphs and Graph Transformer Networks}
\begin{figure*}[t]
    \centering
    \includegraphics[width=16cm, height=6cm]{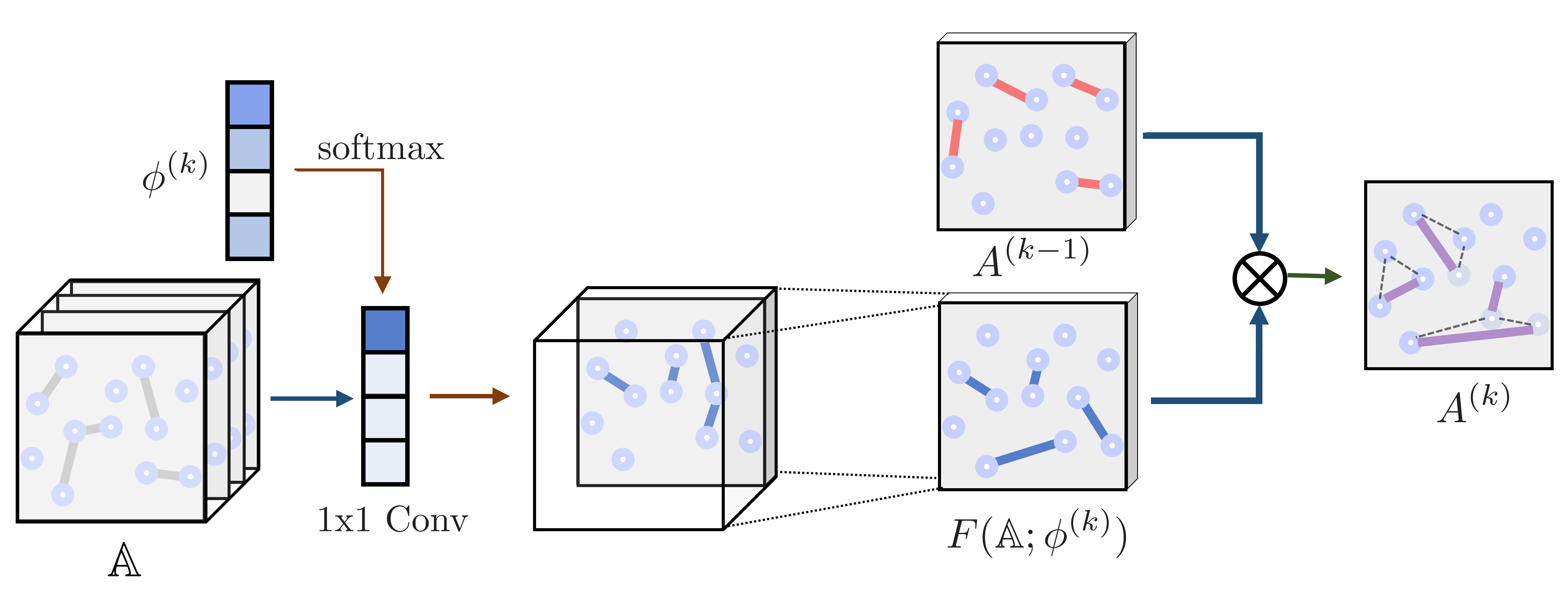}
    \caption{\textbf{Graph Transformer Layer} softly selects adjacency matrices (edge types) from the set of adjacency matrices $\mathbb{A}$ of a heterogeneous graph $G$ and \tpami{learns a new meta-path graph represented by $A^{(k)}$ via the matrix multiplication of the output matrix of the previous $(k-1)$-th GT Layer and the selected adjacency matrix $F(\mathbb{A};\phi^{(k)})$.} The soft adjacency matrix selection is a weighted sum of candidate adjacency matrices obtained by $1\times1$ convolution with non-negative weights from softmax($\phi^{(k)}$).}
    \label{fig:GTLayer}
\end{figure*}
\subsection{Preliminaries}
 \textbf{Heterogeneous Graph \cite{shi2016survey}.} 
Let $\mathcal{G}=(\mathcal{V},\mathcal{E},\mathcal{T}_{v}, \mathcal{T}_e)$ denote a directed graph where each node v $\in \mathcal{V}$ and each edge e $\in \mathcal{E}$ are associated with their type mapping functions $\tau_v{(v)}: \mathcal{V} \rightarrow \mathcal{T}_v$ and $\tau_e{(e)}: \mathcal{E} \rightarrow \mathcal{T}_e$, respectively.
 The heterogeneous graph $\mathcal{G}$ can be represented by a set of adjacency matrices $\{A_t\}_{t=1}^{|\mathcal{T}_e|}$ or a tensor (i.e., $\mathbb{A} \in \Rb^{|V| \times |V| \times |\mathcal{T}_e|}$), where $A_t \in \Rb^{N \times N}$ is an adjacency matrix of the $t$-th edge type and $|V| = N$. $A_t[i,j]$ denotes the weight of the $t$-th type edge from node $j$ to node $i$. When a graph has a single type of nodes and edges, i.e., $|\mathcal{T}_v|=1$ and $|\mathcal{T}_e|=1$, it is called a \textit{homogeneous} graph.

\textbf{Meta-Path \cite{wang2019han}}.  In heterogeneous graphs, a multi-hop connection is called a \textit{meta-path}, which is a path connected with heterogeneous edge types, i.e., ${v}_1 \xrightarrow{\tau_e{(e_1)}} {v}_2 \xrightarrow{\tau_e{(e_2)}} \ldots  \xrightarrow{\tau_e{(e_\ell)}} v_{\ell+1}$, where $\tau_e{(e_\ell)} \in \mathcal{T}_e$ denotes the edge type of edge $e_\ell$ on the meta-path.

\textbf{Graph Convolutional network (GCN).}
In this work, a graph convolutional network (GCN)~\cite{kipf2016semi} is used to learn useful representations for node classification in an end-to-end fashion. % on learned graph structures. 
Let $H^{(l)}$ be the feature representations of the $l$th layer in GCNs, the forward propagation becomes
\begin{equation}
{H}^{(l+1)} = \sigma \left (\Tilde{D}^{-\frac{1}{2}}\Tilde{A}\Tilde{D}^{-\frac{1}{2}}H^{(l)}W^{(l)} \right ), 
\label{eq:gcn}
\end{equation}
where $\Tilde{A} = A+I \in \Rb^{N\times N}$ is the adjacency matrix $A$ of the graph $G$ with added self-connections, $\tilde{D}$ is the degree matrix of $\tilde{A}$, i.e., $\Tilde{D}_{ii}=\sum_{i}{\Tilde{A}_{ij}}$, and $W^{(l)} \in \Rb^{d \times d}$ is a trainable weight matrix. One can easily observe that the convolution operation across the graph is determined by the given graph structure and it is not learnable except for the node-wise linear transform $H^{(l)} W^{(l)}$. So the convolution layer can be interpreted as the composition of a fixed convolution followed by an activation function $\sigma$ on the graph after a node-wise linear transformation. 
Since we learn graph structures, our framework benefits from the different convolutions, namely $\Tilde{D}^{-\frac{1}{2}}\Tilde{A}\Tilde{D}^{-\frac{1}{2}}$, obtained from learned multiple adjacency matrices. The architecture will be introduced later in this section.
For a directed graph (i.e., asymmetric adjacency matrix), $\tilde{A}$ in ~\eqref{eq:gcn} can be normalized by the inverse of in-degree diagonal matrix $D^{-1}$ as ${H}^{(l+1)} = \sigma(\Tilde{D}^{-1}\Tilde{A}H^{(l)}W^{(l)})$.

\begin{figure*}[t]
    \centering
    \includegraphics[width=18cm, height=8.5cm]{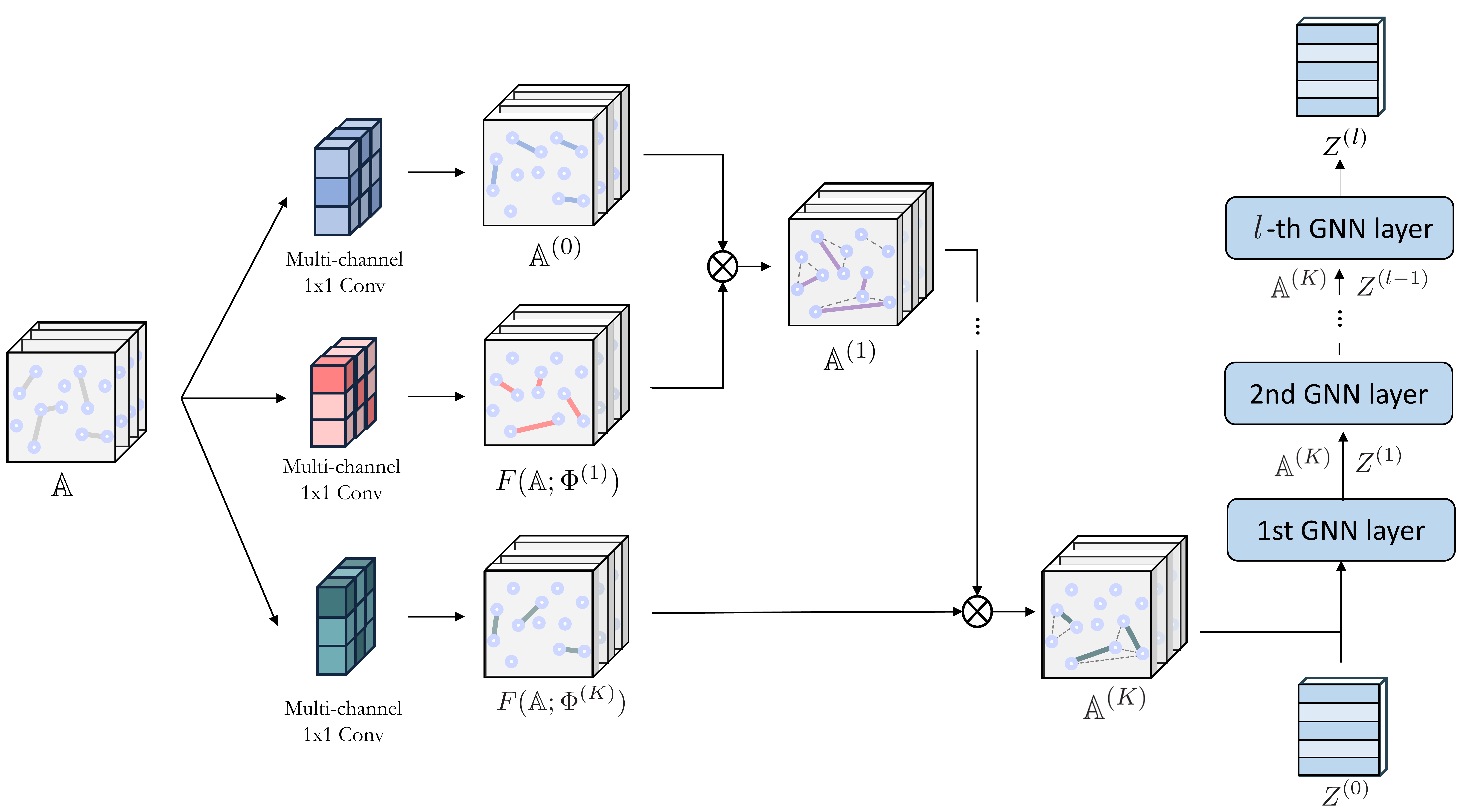}
    \caption{Graph Transformer Networks (GTNs) learn to generate a set of new meta-path adjacency matrices $\mathbb{A}^{(K)}$ using GT layers and perform graph convolution as in GCNs on the new graph structures. Multiple node representations from the same GCNs on multiple meta-path graphs are integrated by concatenation and improve the performance of node classification. 
    % $\mathbb{Q}_1^{(l)}$ and $\mathbb{Q}_2^{(l)} \in \Rb^{N \times N \times C}$ are intermediate
    $F(\mathbb{A};\Phi^{(K)})$ is an intermediate adjacency tensor to compute meta-paths at the $K$th layer.}
    \label{fig:GTN}
\end{figure*}
\begin{figure*}[t]
    \centering
    \includegraphics[width=18cm, height=8.5cm]{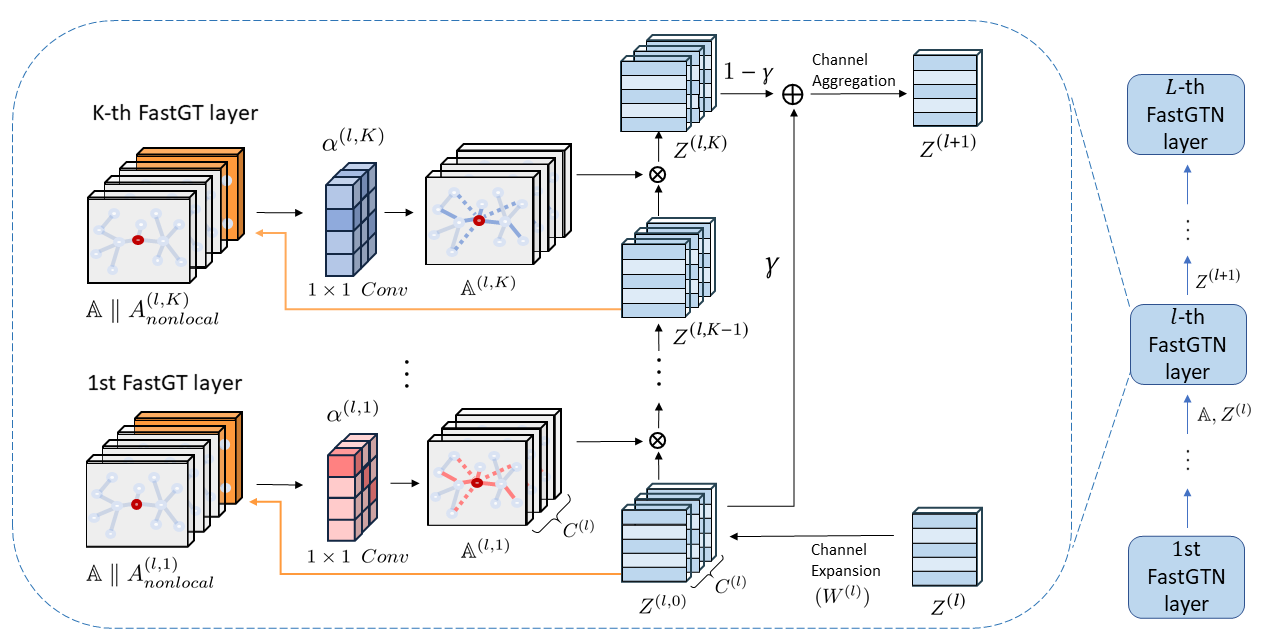}
    \caption{Fast Graph Transformer Networks (FastGTNs) implicitly transform graph structures by a sequence of feature transformations using differently constructed adjacency matrices from Fast Graph Transformer Layers (FastGT Layers). 
    Each $k$-th FastGT Layer first generates a non-local adjacency matrix $A_{non local}^{(l,k)}$ using the hidden representations $Z^{(l, k-1)}$ from $(k-1)$-th FastGT layer and appends it to the set of adjacency matrices $\mathbb{A}$. 
    To generate diverse graphs, $k$-th FastGT layer generates $C^{(l)}$ new adjacency matrices $\mathbb{A}^{(l,k)} \in \Rb^{N \times N \times C^{(l)}}$ by applying a $1 \times 1$ convolution filter.
    Using the $C^{(l)}$ new adjacency matrices in $\mathbb{A}^{(l,k)}$, a FastGT layer transforms the hidden presentations $Z^{(l, k-1)}$ into $Z^{(l, k)}$. 
    After $K$ FastGT layers, the final representations are obtained by channel aggregation after a convex combination of the expanded input representations $Z^{(l,0)}$ and the output representations of the $K$-th FastGT layer, i.e., $Z^{(l+1)} = f_{agg}(\gamma Z^{(l,0)}+(1-\gamma) Z^{(l,K)})$, where $\gamma \in (0,1)$.
    }
    \label{fig:FastGTNs_model}
    % Fast Graph Transformer Networks (FastGTNs) learn to transform node representations $Z^{(l)}$ using differently constructed adjacency matrices $\mathbb{A}^{(k)}$ from each FastGT layer. To consider semantic proximity between nodes, $k$-th FastGT layer first generates a non-local graph $A_{non local}^{(k)}$ using the hidden representations of $(k-1)$-th FastGT layer and append $A_{non local}^{(k)}$ to the set of adjacency matrices $\mathbb{A}$. Then given the set of adjacency matrices $\mathbb{A}$, $k$-th FastGT layer transform the previous hidden representations using $C$ softly selected adjacency matrices $A^{(k)}$ by 1 x 1 convolution. After stack of $K$ FastGT layers, the final representations are obtained by weighted sum of the input node representations $Z^{(l)}$ and the output representations of the $K$-th FastGT layer.
\end{figure*}
\subsection{Learning Meta-Path Graphs}
\label{sub:Learning Meta-Path Graphs}
Previous works \cite{wang2019han, zhang2018deep} require manually defined meta-paths and perform Graph Neural Networks on the meta-path graphs. 
Instead, our Graph Transformer Networks (GTNs) learn meta-path graphs for given data and tasks and operate graph convolution on the learned meta-path graphs. 
This gives a chance to find more useful meta-paths and lead to virtually various graph convolutions using multiple meta-path graphs. 

\tpami{The main idea of learning meta-path graphs is that a new adjacency matrix $A_\mathcal{P}$ of a useful meta-path $\mathcal{P}$ connected by a particular order of edge types (e.g., $t_1, t_2 \ldots  t_\ell$) is obtained by the multiplications of adjacency matrices of the edge types  as
\begin{equation}
{A}_\mathcal{P} = A_{t_\ell} \ldots  A_{t_2} A_{t_1}.
\label{eq:metapath}
\end{equation}
Based on this idea, each $k$-th Graph Transformer (GT) Layer in GTN learns to softly select adjacency matrices (edge types) by $1\times 1$ convolution with the weights from softmax function as \begin{align}
F(\mathbb{A};\phi^{(k)}) &= \text{conv}_{1\times1}(\mathbb{A};\text{softmax}(\phi^{(k)})) \\
                         &= \sum_{t=1}^{|\mathcal{T}_e|} \alpha_t^{(k)} A_t,
\end{align}
where $\phi^{(k)} \in \Rb^{1 \times 1 \times |\mathcal{T}_e|}$ is the parameter of $1\times 1$ convolution and  $\alpha^{(k)}=\text{softmax}(\phi^{(k)})$ (i.e., the convex combination of adjacency matrices as $\alphab^{(k)} \cdot \mathbb{A}$).
Then the meta-path adjacency matrix is computed by matrix multiplication of an output and the output matrix of the previous $(k-1)$-th GT Layer as $A^{(k-1)}F(\mathbb{A};\phi^{(k)})$. For numerical stability, the matrix is normalized by its degree matrix as
\begin{equation}
 \label{eq:att_score}
{A}^{(k)} = {\left (\hat{D}^{(k)} \right )^{-1}}A^{(k-1)}F(\mathbb{A};\phi^{(k)}) ,
\end{equation}
where ${A}^{(0)} = F(\mathbb{A};\phi^{(0)})$ and $\hat{D}^{(k)}$ is a degree matrix of the output after the multiplication of two matrices $A^{(k-1)}F(\mathbb{A};\phi^{(k)})$.}

Now, we need to check whether GTN can learn an arbitrary meta-path with respect to edge types and path length. 
%Let $\mathbb{A} \in \Rb^{N \times N \times K}$ be a set of adjacency matrices for a heterogeneous graph $G$, where $N$ and $K$ are the number of nodes and edge types respectively. 
The adjacency matrix of arbitrary length $k+1$ meta-paths can be calculated by 
% We first reformulate Eqn. \eqref{eq:metapath} to consider all possible meta-paths in graph as
%\begin{equation}
%{A}_{P} = \left(\sum_{i=1}^{K}{\alpha_{i}^{(1)} A_i}\right) \left(\sum_{i=1}^{K}{\alpha_{i}^{(2)} A_i}\right) \ldots \left(\sum_{i=1}^{K}{\alpha_{i}^{(l)} A_i}\right)
%\label{eq:weighted_sum}
%\end{equation}

\begin{align}
{A}_{P} = \left(\sum_{t_0 \in \mathcal{T}^e}{\alpha_{t_0}^{(0)} A_{t_0}}\right) \left(\sum_{t_1 \in \mathcal{T}^e}{\alpha_{t_1}^{(1)} A_{t_1}}\right) \ldots \left(\sum_{t_{k} \in \mathcal{T}^e}{\alpha_{t_{k}}^{(k)} A_{t_{k}}}\right)
\label{eq:weighted_sum}
\end{align}
where $A_{P}$ denotes the adjacency matrix of meta-paths, $\mathcal{T}^e$ denotes a set of edge types and $\alpha_{t_l}^{(k)}$ is the weight for edge type $t_k$ at the $k$-th GT layer. When $\alpha$ is not \sjy{one-hot vector},  ${A}_{P}$ can be seen as the weighted sum of all length-($k+1$) meta-path adjacency matrices. So a stack of $k$ GT layers allows to learn length $k+1$ meta-path graph structures as the architecture of GTN shown in Fig. \ref{fig:GTN}.
%is the weight of $l$-th adjacency matrix $A_i$ for the edge type $i$
One issue with this construction is that adding GT layers always increase the length of meta-path and this does not allow the original edges. In some applications, both long meta-paths and short meta-paths are important. 
To learn short and long meta-paths including original edges, we include the identity matrix $I$ in $\mathbb{A}$, i.e., $A_{0}=I$. This trick allows GTNs to learn any length of meta-paths up to $k+1$ when $k$ GT layers are stacked.

\subsection{Graph Transformer Networks}\label{method-graph_transformer}
We here introduce the architecture of Graph Transformer Networks. To consider multiple types of meta-paths simultaneously, the GTN generates multiple graph structures by setting the output channels of $1 \times 1$ filter to C. \tpami{Then the output matrix of $k$-th GT layer $A^{(k)}$ becomes the output tensor $\mathbb{A}^{(k)} \in \Rb^{N \times N \times C}$ and the weight vector $\phi^{(k)}$ of the $k$-th GT Layer becomes the weight matrix $\Phi^{(k)}$. Eq (\ref{eq:att_score}). can be represented in the form of tensor equations as
\begin{equation}
\mathbb{A}^{(k)} = {\left (\hat{\mathbb{D}}^{(k)} \right )^{-1}}\mathbb{A}^{(k-1)} * F(\mathbb{A};\Phi^{(k)}),
\end{equation}
where $\mathbb{A}^{(k-1)} * F(\mathbb{A};\Phi^{(k)}) = \concat{c}{C}{A^{(k-1)}_c F(\mathbb{A};\phi^{(k,c)})}$ and $\hat{\mathbb{D}}^{(k)}$ is a degree tensor of the output after the multiplication of two tensors $\mathbb{A}^{(k-1)} * F(\mathbb{A};\Phi^{(k)})$. It is beneficial to learn different node representations via multiple different graph structures.}

\tpami{After the stack of $K$ GT Layers, multi-layer GNNs are applied to the each channel of the output tensor $\mathbb{A}^{(K)}$ and update node representations $Z$ as follows:
\begin{equation}
Z^{(l+1)} = f_{agg}\left (\concat{c}{C} \sigma(\Tilde{D}^{-1}_c \tilde{A}_{c}^{(\icml{K})}Z^{(l)}W^{(l)})\right ),
\label{eq:GTN}
\end{equation}
where \text{\Large{\textdoublevertline}} is the concatenation operator, C denotes the number of channels, $Z^{(l)}$ denotes the node representations at the $l$-th GNN layer, $\tilde{A}^{(K)}_c = A^{(K)}_c+\gamma I $ is the adjacency matrix with self-loops from the $c$-th channel of $\mathbb{A}^{(K)}$, $\tilde{D}_c$ is the degree matrix of $\tilde{A}^{(K)}_c$, $W^{(l)} \in \Rb^{d^{(l)} \times d^{(l+1)}}$ is a trainable weight matrix shared across channels, $Z^{(0)}$ is a feature matrix $X \in \Rb^{N \times F}$ and $f_{agg}$ is a channel aggregation function.}

The final node representations $Z^{(l)}$ after $l$ GNN layers can then be used for downstream tasks. For the node classification task, we applied dense layers followed by a softmax layer to the node representations. Then with ground truth labels of nodes, we can optimize the model weights by minimizing the cross-entropy via backpropagation and gradient descent.
% %\begin{equation}
% %\Bar{y} = \text{MLP}(ReLU(\text{MLP}(\Tilde{H}))
% %\end{equation}
% \begin{equation}
% \mathcal{L} = \frac{1}{n}\sum_{i=1}^{n}{\mathcal{L}_c(y_i,C \cdot Z_i}),
% \end{equation}
% where $n$ denotes the number of labels,$\mathcal{L}_c$ is the classification loss (\eg, the cross entropy), C is the parameter of the classifier, $y_i$ and $Z_i$ are the labels and embeddings of labeled nodes.
% We optimize the proposed model via back propagation with labeled data and learn the embeddings of nodes.
\subsection{Fast Graph Transformer Networks}
\label{sec:3.2}
\hjk{In the previous sections, we demonstrated that GTNs can transform the original graphs into new meta-path graphs while learning representations on the meta-path graphs. However, GTNs have a scalability issue. GTNs \textit{explicitly} compute a new adjacency matrix of meta-paths by the matrix multiplication of two adjacency matrices and store the new adjacency matrix at each layer.
So a graph transformation in GTNs involves huge computational costs and large memory.} 
\tpami{This makes it infeasible to apply GTNs to a large graph. 
To address these issues, we develop the enhanced version of GTNs, FastGTNs, which \textit{implicitly} transform the graph structures without storing the new adjacency matrices of meta-paths.
In this section, we describe our FastGTNs in detail.}

\tpami{To derive FastGTNs, we first begin with the equation of GTNs.
Our goal here is to derive a new architecture without the need for the explicit multiplications of large adjacency matrices.
For simplicity, we assume that the number of channels is one, i.e., $C$=1, one GCN Layer is applied on top of the new graph structure and a channel aggregation function is an identity function i.e., $f_{agg}(x)=x$.
Then the node representations $Z$ of the GTNs are given as follows:
\begin{align}
Z &= \sigma(\Tilde{D}^{-1} (A^{(K)}+I)XW), 
\label{eq:GTNstart}
\end{align}
where $A^{(K)}\in \Rb^{N \times N}$ is a new adjacency matrix from a GTN with $K$ GT layers, and $X\in \Rb^{N \times F}$ is input features, $W\in \Rb^{F \times d'}$ is a linear transformation in a GCN layer, $\Tilde{D}^{-1}\in \Rb^{N \times N}$ is an inverse degree matrix of ($A^{(K)}+I$).
We observe that a GT Layer multiplies two softly selected adjacency matrices and normalizes the output adjacency matrix. So \eqref{eq:GTNstart} can be written as
\begin{align}
    \begin{split}
        Z &= \sigma(\Tilde{D}^{-1}XW + \Tilde{D}^{-1}A^{(K)}XW) \\
        &= \sigma (\Tilde{D}^{-1}XW + \Tilde{D}^{-1} \left (\invhatD{K} A^{(K-1)} (\alphab^{(K)} \cdot \mathbb{A}) \right) \\
        &\phantom{{}={}} XW ) \\
        &= \sigma ( \Tilde{D}^{-1}XW +  \Tilde{D}^{-1} \bigg( \invhatD{K} \dots \bigg( \invhatD{1} \\
        &\phantom{{}={}} (\alphab^{(0)} \cdot \mathbb{A}) (\alphab^{(1)} \cdot \mathbb{A}) \bigg) \dots (\alphab^{(K)} \cdot \mathbb{A}) \bigg) XW ).
    \end{split}
\label{eq:gtn_Z}
\end{align}
The Equation (\ref{eq:gtn_Z}) clearly shows the computational bottleneck in (\ref{eq:gtn_Z}) is the multiplications of huge adjacency matrices, e.g., $(\alphab^{(0)} \cdot \mathbb{A}) (\alphab^{(1)} \cdot \mathbb{A}) \dots (\alphab^{(K)} \cdot \mathbb{A})$.
To resolve this problem, we can rewrite (\ref{eq:gtn_Z}) using the associative property of matrix multiplication as
\begin{align}
\label{eq:associative}
Z &= \sigma(\Tilde{D}^{-1}XW + \Tilde{D}^{-1}\invhatD{K} \dots \invhatD{1}  \\
&\phantom{{}={}} (\alphab^{(0)} \cdot \mathbb{A} (\alphab^{(1)}\cdot \mathbb{A} \nonumber \dots (\alphab^{(K)} \cdot \mathbb{A}XW)))). 
\end{align} }

\tpami{Now, Equation \eqref{eq:associative} implies that at each layer, without the \textit{matrix multiplications} of huge adjacency matrices, the identical features can be obtained by a sequence of \textit{feature} transformations using a differently constructed adjacency matrix, e.g., $\alphab^{(k)} \cdot \mathbb{A}H_{k}$ .
It efficiently reduces the computational cost from $O(N^3)$ to $O(N^2F)$ and memory usage from $O(N^2)$ to $O(NF)$. However, note that since we do not compute the multiplication of two adjacency matrices anymore, we cannot compute degree matrices $\Tilde{D}^{-1}\invhatD{K} \dots \invhatD{1}$.}

\tpami{To address this challenge, now we show that given a condition of input data and a proposition of normalized matrices, we can make all degree matrices in (\ref{eq:associative}) into identity matrices. Then we can compute (\ref{eq:associative}) without the matrix multiplications of huge adjacency matrices. We begin with a proposition and a condition, then derive an equation of our FastGTNs from (\ref{eq:associative}) by using the Proposition \ref{prop:normalized}.
\begin{proposition}
Given two normalized adjacency matrices $A$, $B \in \Rb^{N \times N}$, the followings are equivalent:
\begin{enumerate}[label=(\roman*)]
    \item $\left( {D}_{A}^{-1}{A}\right) \left({D}_{B}^{-1}{B}\right) = \left({D}_{AB}^{-1}{AB} \right)$
    \item ${D}_{AB}^{-1} = I$
    \item ${D}_{A+I}^{-1}=({D}_{A}+I)^{-1}=\cfrac{1}{2}~I$
\end{enumerate}
\label{prop:normalized}
\end{proposition}
The proof of Proposition \ref{prop:normalized} is provided in A.2 in the supplement. We first assume that each adjacency matrix in $\mathbb{A}$ is row-wise normalized i.e., $\sum\limits_{j} A_{t}[i,j]=1$. 
The convex combination of adjacency matrices at each $k$-th layer i.e., $\alphab^{(k)} \cdot \mathbb{A}$ is also a normalized matrix. 
Then $\invhatD{k}$ is an inverse degree matrix of the output after multiplication of two normalized matrices $A^{(k-1)}(\alphab^{(K)} \cdot \mathbb{A})$. It means that by (\romannumeral 2) in Proposition 1, all $\invhatD{k}$ at each $k$-th layer are the identity matrix $I$ and, thus, (\ref{eq:associative}) can be rewritten as 
\begin{align}
Z &= \sigma(\Tilde{D}^{-1}XW + \Tilde{D}^{-1}(\alphab^{(0)} \cdot \mathbb{A} (\alphab^{(1)}\cdot \mathbb{A} \dots (\alphab^{(K)} \\
&\phantom{{}={}} \cdot \mathbb{A}XW)))). \nonumber
\end{align}
By (\romannumeral 3) in Proposition 1, we can also know that $\tilde{D}^{-1}=(D_{A^{(K)}} + I)^{-1} = \frac{1}{2} I$, then $Z$ can be represented as
\begin{align}
Z &= \sigma(\frac{1}{2}XW+\frac{1}{2}(\alphab^{(0)} \cdot \mathbb{A} (\alphab^{(1)}\cdot \mathbb{A} \dots (\alphab^{(K)} \cdot \mathbb{A}XW)))).
\end{align}
Since each layer constructs one convex combination of adjacency matrices, K-layers generate K-adjacency matrices as
\begin{align}
Z &= \sigma(\cfrac{1}{2} XW+ \cfrac{1}{2}(\alphab^{(1)} \cdot \mathbb{A} (\alphab^{(2)}\cdot \mathbb{A} \dots (\alphab^{(K)} \cdot \mathbb{A}XW)))).
\end{align}
Now this derivation means that our FastGTNs are not an approximation of GTNs. Mathematically, they are exactly identical. We'll discuss more in Section \ref{sub:identity btw GTN and FastGTN} about the identity.
We reverse the order of layers from 1 to K and replace $\cfrac{1}{2}$ with a hyper-parameter $\gamma$, then the output of a FastGTN can be represented as
\begin{align}
Z &= \sigma(\gamma XW+ (1-\gamma)(\alphab^{(K)} \cdot \mathbb{A}  \dots
 (\alphab^{(1)} \cdot \mathbb{A}XW)))). 
\label{eq:FastGTN}
\end{align}
We finally rewrite our FastGTNs for multi-channel and multi-layer settings as
\begin{align}
Z^{(l+1)} &= f_{agg}\bigg(~\concat{c}{C^{(l)}} \sigma(\gamma Z^{(l)}W^{(l)}_c + 
                                 (1-\gamma)Z_c^{(l,K)}) \bigg),
\label{eq:FastGTN_multi_layer}
\end{align}
\begin{align}
Z_c^{(l,K)} &= (\alphab^{(l,K,c)} \cdot \mathbb{A} \dots (\alphab^{(l,1,c)} \cdot \mathbb{A}Z^{(l)}W^{(l)}_c)),
\label{eq:FastGTN_multi_layer1}
\end{align}
where $C^{(l)}$ denotes the number of channels, $Z^{(l)}$ denotes the node representations from the $l$-th FastGTN layer, $W^{(l)}_c\in \Rb^{d^{(l)} \times d^{(l+1)}}$ is a linear transformation in $c$-th channel of the $l$-th FastGTN layer, $\mathbb{A}$ is a set of normalized adjacency matrices, $\alphab^{(l,k,c)}$ is a convolution filter in the $c$-th channel of the $k$-th FastGT layer in the $l$-th FastGTN layer, $Z^{(0)}$ is a feature matrix $X\in \Rb^{N \times F}$ and $f_{agg}$ is a channel aggregation function. }
\hjk{Furthermore, to deal with huge graphs with about 30 million edges, we additionally propose a mini-batch training algorithm for GTNs and FastGTNs in A.1 in the supplement.}

\subsection{Non-Local Operations.}
\hjk{One limitation of the GTNs is that its transformation is limited to compositions of existing relations. 
Specifically, K GT layers can generate edges only up to (K+1)-hop relations. 
It cannot generate remote relations based on semantic proximity between nodes.
To address this limitation, we extend graph transformations to non-local operations incorporating the node features to utilize the semantic proximity of nodes beyond meta-paths. 
However, as mentioned in the previous section, since GTN itself requires large computation cost, we extend non-local operations only to FastGTNs.}
\tpami{Specifically, at each $k$-th FastGT layer in each $l$-th FastGTN layer, we construct a non-local adjacency matrix $A_{\text{non local}}^{(l, k)} \in \Rb^{N \times N}$ based on hidden representations  $Z^{(l,k-1)}$ from the previous FastGT layer and append the non-local adjacency matrix to the candidate set of adjacency matrices $\mathbb{A}$ to utilize the non-local relations for graph transformations. To construct $A_{\text{non local}}^{(l,k)}$, we first calculate a graph affinity matrix at each $k$-th FastGT Layer $M^{(l,k)}$ based on the similarity between node features. We take an average of multi-channel hidden representations from each $(k-1)$-th FastGT Layer and project the averaged representations into a latent space by a non-linear transformation $g_{\theta}$. Then we compute the affinity matrix $M^{(l,k)}$ using the similarity in the latent space as
\begin{equation}
M^{(l,k)} = (g_{\theta}(H^{(l,k-1)})g_{\theta}(H^{(l,k-1)})^T),    
\end{equation}
\begin{equation}
H^{(l,k-1)} = \frac{1}{C^{(l)}}\sum_{c=1}^{C^{(l)}}{Z^{(l,k-1)}_c},    
\end{equation}
where $Z^{(l,k-1)}$ denotes hidden representations at a $(k-1)$-th FastGT Layer.
We use the trick of a decoder in GAE \cite{kipf2016variational} as the similarity function to get the affinity matrix.
The affinity matrix can be seen as a weighted adjacency matrix of the fully connected graph. If we include the dense affinity matrix as an adjacency matrix for graph transformation, it causes huge computation cost and may rather propagate irrelevant information between nodes. Therefore, we sparsify the affinity matrix by extracting only $n$ largest weights for each node $i$ and construct non-local adjacency matrix $A_{\text{non local}}^{(l,k)}$ as
\begin{gather}
    A_{\text{non local}}^{(l,k)}[i,j] = 
    \begin{cases}
    M_{ij}^{(l,k)}, & \mbox{if } j \in arg\,top\,k(M^{(l,k)}[i,:], n) \\
    0, & \mbox{otherwise }
    \end{cases}
\end{gather}
We row-wise normalize the non-local adjacency matrix by applying the softmax function to edge weights of each row. 
Then the final non-local adjacency matrix at each $k$-th FastGT layer is represented as
\begin{equation}
    A_{\text{non local}}^{(l,k)}[i,:] = \text{softmax}(topk(M^{(l,k)}[i,:], n)).
\end{equation}
To use the normalized non-local adjacency matrix for transformations, we add a non-local parameter to 1x1 convolution filters of each FastGT Layer and then append the matrix $A_{\text{non local}}^{(l,k)}$ to the set of adjacency matrix of $k$-th FastGT Layer i.e., $ \mathbb{A} \text{\Large{\textdoublevertline}} A_{\text{non local}}^{(l,k)}$}.
\subsection{Relations to Other GNN Architectures}
\label{sub:relations to other GNN architectures}
\hjk{FastGTNs enhance the scalability of GTNs by implicitly transforming graph structures. Moreover, the FastGTNs become a flexible/general model that subsumes other graph neural networks.
In this section, we discuss relationships between our FastGTNs and other GNN architectures.} \tpami{Interestingly, if input graphs are normalized i.e., $D=I$, several popular graph neural networks such as GCN \cite{kipf2016gcn} and MixHop \cite{abu2019mixhop} can be special cases of our FastGTNs. In addition, RGCN \cite{schlichtkrull2018rgcn} can be subsumed by our FastGTNs with minor modifications.
We first discuss the graph convolution network (GCN). The GCN computes the output node representations from the $l$-th GCN layer as 
\begin{align}
{Z}^{(l+1)} &= \sigma \left (\Tilde{D}^{-\frac{1}{2}}\Tilde{A}\Tilde{D}^{-\frac{1}{2}}Z^{(l)}W^{(l)} \right )\\
            &= \sigma \left (\frac{1}{2}Z^{(l)}W^{(l)}+\frac{1}{2}AZ^{(l)}W^{(l)} \right ),
\label{eq:gcn2}
\end{align}
where $\Tilde{A} = A+I \in \Rb^{N\times N}$ and $\tilde{D}$ is the degree matrix of $\tilde{A}$. 
If the number of FastGT layers in our FastGTNs is one i.e., $K=1$, the number of channels is one, i.e., $C=1$ and $\gamma$ equals $\frac{1}{2}$, the output node representations are as 
\begin{align}
Z^{(l+1)} &= \sigma(\frac{1}{2}Z^{(l)}W^{(l)}+\frac{1}{2}(\alphab^{(l,1)}\cdot \mathbb{A})Z^{(l)}W^{(l)}).
\label{eq:FastGTNs_gcn}
\end{align}
Then if the first FastGT layer only selects the adjacency matrix, i.e., $\icml{\alphab^{(l,1)}}\cdot \mathbb{A} = 1 \cdot A+0 \cdot I$, the output representations are exactly same as the output of the GCNs.}

\tpami{Mixhop \cite{abu2019mixhop} is an extended GNN architecture which can capture long-range dependencies by mixing powers of the adjacency matrix as 
\begin{equation}
 Z^{(l+1)} = \underset{j \in P}{{ \text{\Large{\textdoublevertline}}}} \sigma(\hat{A}^{j}Z^{(l)}W^{(l)}_j),
\label{eq:mixhop}
\end{equation}
where $P$ is a set of integer adjacency powers, $\hat{A}$ is a symmetrically normalized adjacency matrix with self-connections, i.e., $\hat{A}=\tilde{D}^{-\frac{1}{2}}(A+I)\tilde{D}^{-\frac{1}{2}}$ and $\hat{A}^{(j)}$ denotes the adjacency matrix $\hat{A}$ multiplied by itself $j$ times. 
Since the degree matrix of $A+I$ equals $2I$, we can rewrite equation~(\ref{eq:mixhop}) as 
\begin{align}
% $\tilde{D}=2I$
%  Z^{(l+1)} &= \underset{j \in P}{{ \text{\Large{\textdoublevertline}}}} \sigma\left(\left(\frac{1}{2}A+\frac{1}{2}I\right)^j Z^{(l)}W^{(l)}_j\right) \\
 Z^{(l+1)} &= \underset{j \in P}{{ \text{\Large{\textdoublevertline}}}} \sigma\left(\left(\alpha \cdot \mathbb{A}\right)^j Z^{(l)}W^{(l)}_j\right),
\label{eq:FastGTN MixHop}
\end{align}
where $\alpha \cdot \mathbb{A} = \frac{1}{2} A + \frac{1}{2} I$.
% If the $\mathbb{A}=\{A, I\}$ and  = we represent $\frac{1}{2}A+\frac{1}{2}I$ as $\alpha \cdot \mathbb{A}=$
Then if $\gamma$ equals $0$, the number of channels equals the size of $P$ i.e., $C=|P|$, number of FastGT layers in each channel equals $j$ i.e., K=j, all FastGT layers choose the adjacency matrix and the identity matrix in the same ratio, i.e., $\alphab^{(l,k)}\cdot \mathbb{A} = \frac{1}{2}A+ \frac{1}{2}I$ and a channel aggregation function is an identity function i.e., $f_{agg}(x)  =x$, the output is same as the output of the Mixhop, i.e., the Mixhop can be a special case of our FastGTNs.}

\tpami{Lastly, we discuss RGCN \cite{schlichtkrull2018rgcn} which extends the GCN to heterogeneous graphs by utilizing relation-specific parameters. Specifically, the output representations from the $l$-th RGCN are as 
\begin{align}
    Z^{(l+1)} &= \sigma{\left(\sum_{t=1}^{|\mathcal{T}_e|}{D^{-1}_{t}A_t Z^{(l)} \left(\sum_{i=1}^B{a_{ti}^{(l)}V_i^{(l)}}\right)}\right)} \label{eq:rgcn}\\
              &= \sigma{\left(\sum_{i=1}^B a_{i}^{(l)} \cdot {\mathbb{A} Z^{(l)} {V_i^{(l)}}}\right)}~, \label{eq:rgcn2}
\end{align}
where $A_t$ denotes an adjacency matrix for relation type $t$, $V_i^{(l)}$ denotes a basis parameter of $l$-th RGCN layer and $a_{ti}^{(l)}$ denotes a coefficient for relation type $t$. 
The derivation from (\ref{eq:rgcn}) to (\ref{eq:rgcn2}) is provided in A.3 in the supplement.
If the number of FastGT layers is one i.e., $K=1$, the number of channels equals number of basis matrices i.e., $C=B$, $\gamma$ equals $0$ and a channel aggregation function $f_{agg}$ is summation, the output node representations are given as
\begin{align}
Z^{(l+1)} &= \sigma\left(\sum_{c}^{B}{\alphab^{(l,1,c)} \cdot \mathbb{A}Z^{(l)}W^{(l)}_c}\right).
\label{eq:RGCN_FastGTN}
\end{align}
Then (\ref{eq:rgcn2}) and (\ref{eq:RGCN_FastGTN}) are exactly same except for that (\ref{eq:rgcn2}) apply \textit{different} \textit{linear} combinations in each layer.}
\begin{table*}[t]
\centering
\begin{tabular}{lcccccccc}
\toprule
\multicolumn{1}{c}{Type} & Dataset   & \# Nodes & \# Edges & \# Features & \# Classes & \# Training & \# Validation & \# Test \\ \midrule
\multirow{6}{*}{Heterogeneous} 
 & DBLP                         & 18405    & 67496    & 334         & 4          & 800         & 400           & 2857    \\ 
 & ACM                           & 8994     & 25922    & 1902        & 3          & 600         & 300           & 2125    \\ 
 & IMDB                          & 12772    & 37288    & 1256        & 3          & 300         & 300           & 2339    \\ 
 & CS                          & 1116163    & 28427508    & 768        & 3505          & 147769         & 33582           & 46711    \\ 
 & ML                          & 227144    & 4249598    & 768        & 1447          & 19902         & 6112           & 7911    \\ 
 & NN                           & 66211    & 845916    & 768        & 929          & 4745         & 1057           & 2311    \\ \midrule
 \multirow{6}{*}{Homogeneous} & \textsc{Air-USA}   & 1190     & 13599    & 238         & 4          & 119         & 238           & 833     \\ 
 & \textsc{Blogcatalog}                      & 5196     & 171743   & 8189        & 6          & 519         & 1039          & 3638    \\ 
 & \textsc{Citeseer}                         & 3327     & 4552     & 3703        & 6          & 120         & 500           & 1000    \\ 
 &  \textsc{Cora}                     & 2708     & 5278     & 1433        & 7          & 140         & 500           & 1000    \\ 
 & \textsc{Flickr}                    & 7575     & 239738   & 12047       & 9          & 757         & 1515          & 5303    \\ 
 & PPI                             & 10076    & 157213   & 50          & 121        & 1007        & 2015          & 7054    \\ \bottomrule
\end{tabular}
\vspace{1mm}
\caption{Statistics of both homogeneous and heterogeneous graph datasets.}
\label{tab:data statistics}
\end{table*}
\section{Experiments}

In this section, we evaluate our proposed methods on both homogeneous and heterogeneous graph datasets. The experiments aim to address the following research questions:

\begin{itemize}
    % \item \textbf{Q1.} Are the new graph structures generated by GTNs effective for learning node representations?
    \item \textbf{Q1}. How effective are the GTNs and FastGTNs with non-local operations compared to state-of-the-art GNNs on both \textit{homogeneous} and \textit{heterogeneous} graphs in node classification?
    \item \textbf{Q2}. Can the FastGTNs \textit{efficiently} perform the \textit{identical} graph transformation compared to the GTNs?
    \item \textbf{Q3}. Can GTNs \textit{adaptively} produce a variable length of meta-paths depending on datasets?
    \item \textbf{Q4}. How can we interpret the importance of each meta-path from the adjacency matrix generated by GTNs?
\end{itemize}

\subsection{Experimental Settings}
% Efficieny_Figure
\noindent\textbf{Datasets.}
We evaluate our method on twelve benchmark datasets for node classification including six homogeneous graph datasets and six heterogeneous graph datasets for node classification. Detailed statistics regarding each dataset can be found in Table~\ref{tab:data statistics}. The datasets for each type (i.e., homogeneous or heterogeneous) are as follows:

\medskip

\begin{itemize}
\item \textbf{Heterogeneous Graph Datasets}
    \begin{itemize}
        \item \textsc{DBLP} and \textsc{ACM} are both citation networks. They differ in the sense that \textsc{DBLP} has three types of nodes (paper (P), author (A), conference (C)) and four types of edges (PA, AP, PC, CP). \textsc{ACM} is similar but has has subject (S) as a node type instead of conference (C), with edge types differing accordingly. \textsc{DBLP} and \textsc{ACM} are node classification datasets with author research area and paper category as labels, respectively.
        \item \textsc{IMDB} is a movie network dataset. It contains three types of nodes (movies (M), actors (A), directors (D)) and uses the genres of movies as labels.
        \tpami{\item \textsc{CS}, \textsc{ML}, \textsc{NN} each refer to domain-specific subgraphs from the Open Academic Graph (OAG) \cite{tang2015line}. OAG is a large citation network with ten types of nodes (paper (P), author (A), field ($L_0, L_1, L_2, L_3, L_4, L_5$), venue (V), institute (I)) and ten types of edges (PA, $PL_0 - PL_5$, PV, AI, PP). The task is to predict the venue that each paper is published at.}
    \end{itemize}

    \smallskip
\item \textbf{Homogeneous Graph Datasets}
    \begin{itemize}
        \item \tpami{\textsc{Air-USA} is a dataset made up of graphs representing airport traffic within the US. Each node represents an airport and edges between nodes indicate the existence of commercial flights between the two airports \cite{wu2019net}.}
        \item \tpami{\textsc{BlogCatalog} and \textsc{Flickr} are both social network datasets, with the former being a blogging platform and the latter being an image and video sharing platform. In both dataset, each node represents a user of the online community and the edges correspond to whether or not users are following each other \cite{huang2017label}.}
        \item \tpami{\textsc{Cora} and \textsc{CiteSeer} are citation network datasets. Both datasets are comprised of nodes which represent papers published in various fields and edges which represent citation links \cite{kipf2016semi}.}
        \item \tpami{\textsc{PPI} refers to the protein-protein interaction network dataset. The network's nodes represent a protein structure that contains features corresponding to different gene sets. Edges refer to the relation between such proteins \cite{hamilton2017inductive}.}
    \end{itemize}

\end{itemize}

\medskip

\noindent\textbf{Baselines.}
\tpami{To evaluate the effectiveness of representations learnt by the Graph Transformer Networks in node classification, we compare GTNs with conventional random walk based baselines as well as state-of-the-art GNN based methods. Since homogeneous graph neural networks (e.g., GCN, GAT, JK-Net, MixHop and GCNII) cannot differentially handle the different types of nodes and edges, we apply them after converting the heterogeneous graphs into homogeneous graphs.}

\begin{itemize}
%         \item metapath2vec \cite{metapath2vec} is also a random walk-based network embedding method, but differs from DeepWalk in the sense that it was designed to address heterogeneous graphs. It performs network embedding via meta-path based random walks and utilizes skip-gram with negative sampling.
    \item \textbf{MLP} is a simple baseline model that uses only node features for prediction.
    \item \textbf{Node2Vec} \cite{grover2016node2vec} is a random walk based network embedding method which was originally designed for embedding homogeneous graphs. In heterogeneous graphs, we ignore the heterogeneity of nodes and edges and run DeepWalk on the entire heterogeneous graph.
    \item \textbf{GCN} \cite{kipf2016gcn} utilizes a first-order approximation of the spectral graph filter to aggregate features from neighbors.
    \item \textbf{GAT} \cite{velivckovic2017gat} leverages an attention mechanism to learn the relative weights between the neighborhood nodes.
    \item \textbf{JK-Net} \cite{xu2018jknet} leverages a variable range of neighborhoods by connecting the last layer of the network with all preceding hidden layers.
    \item \textbf{MixHop} \cite{abu2019mixhop} mixes powers of the adjacency matrices and applies a GCN to capture long-range dependencies.
    \item \textbf{GCNII} \cite{chen2020simple} improves GCN with initial residual connection and identity mapping to prevent over-smoothing.
    \item \textbf{RGCN} \cite{schlichtkrull2018rgcn} employs GCNs with relation-specific weight matrices to deal with heterogeneous graphs.
    \item \textbf{HAN} \cite{wang2019han} uses manually selected meta-paths to transform a heterogeneous graph into a homogeneous graph and then applies GNNs on the homogeneous graph. 
    %We select meta-paths in each dataset as described in \cite{wang2019han}.
    \item \textbf{HGT} \cite{hu2020hgt} parameterizes the meta relation triplet of each edge type and uses a structure that utilizes the self-attention of the transformer architecture \cite{vaswani2017attention} to learn specific patterns of different relationships.
    % \item \textbf{GTN} \cite{yun2019graph} learns to transform original graph structures into new graph structures identifying useful meta-paths for given tasks and generate node representations by applying GNNs on the transformed graph structures.
\end{itemize}

\smallskip

\noindent\textbf{Implementation details.}
\tpami{All the models in this paper are implemented using PyTorch and PyTorch Geometric~\cite{fey2019torchgeometric} and the experiments are conducted on a single GPU (Quadro RTX 8000). 
% For random walk based models, a walk length is set to 100 per node for 1000 iterations and the window size is set to 5 with 7 negative samples. 
For Node2Vec, GCN, GAT, JK-Net, GCNII, and RGCN, we used the implementations in PyTorch Geometric. 
We re-implemented HAN and HGT referencing the code from the authors of the papers \cite{wang2019han, hu2020hgt}.
We set the dimensionality of hidden representations to 64 throughout the neural networks and apply the Adam optimizer for all models. 
For each model and each dataset, we perform a hyper-parameter search within the following ranges: the learning rate is from 1e-3 to 1e-6, the dropout rate is from $0.1$ to $0.8$ and the epoch is from $50$ to $200$.%, and the number of GNN layers are from $2$ to $4$.
 Based on the accuracy on validation sets, the best models are selected and the models are used for evaluation.
From ten independent runs, the mean and standard deviation of micro-F1 scores on test datasets are computed.
In our FastGTNs, as discussed in Proposition 1,  we perform the row-wise normalization of the adjacency matrices in $\mathbb{A}$.
To avoid the division-by-zero, we add a small positive number to the diagonal elements of the adjacency matrices, i.e., $D^{-1}(A+\epsilon I)$. }

\renewcommand{\arraystretch}{1.3}
\begin{table*}[h]
\centering
\begin{tabular}{ l c c c c c c }
\toprule
\multicolumn{1}{c}{\multirow{2}{*}{Model}} & \multicolumn{2}{c}{\textsc{CS}} & \multicolumn{2}{c}{\textsc{ML}} & \multicolumn{2}{c}{\textsc{NN}} \\
\cmidrule{2-7}
                      & NDCG       & MRR       & NDCG       & MRR       & NDCG       & MRR       \\
\midrule
MLP               & 28.57±0.005  & 12.34±0.004 & 30.89±0.007  & 14.38±0.007 & 25.99±0.002  & 12.03±0.002 \\
Node2Vec               & 26.92±0.005  & 10.93±0.003 & 31.62±0.001  & 15.04±0.001 & 26.16±0.006  & 11.27±0.007 \\
\midrule
GCN                    & 34.72±0.006  & 17.87±0.005 & 37.76±0.005  & 20.66±0.005 & 31.30±0.002  & 15.89±0.003 \\
GAT                    & 41.40±0.009  & 24.01±0.008 & 37.83±0.012  & 20.87±0.012 & 29.38±0.007  & 14.03±0.009 \\
RGCN                   & 41.83±0.009  & 24.35±0.007 & 39.04±0.004  & 21.71±0.003 & 29.84±0.006  & 14.76±0.007 \\
HGT                    & 42.57±0.017  & 25.00±0.017 & 37.31±0.011  & 20.26±0.009 & 28.43±0.016  & 13.70±0.014 \\
\midrule
$\text{GTN}_{-I}$      & 41.81±0.018  & 24.53±0.017 & \textbf{40.17±0.014}  & \textbf{23.27±0.015} & 31.57±0.006  & 16.33±0.003 \\
GTN                    & \textbf{42.75±0.012}  & \textbf{25.35±0.011} & 39.60±0.016  & 22.70±0.017 & 31.76±0.006  & 16.49±0.008 \\
FastGTN                    & 42.32±0.018  & 24.96±0.019 & 38.65±0.018  & 21.72±0.018 & \textbf{32.02±0.006}  & \textbf{16.72±0.006} \\
\bottomrule
\end{tabular}
\vspace{1mm}
\caption{Node classification performance (NDCG and MRR) on large-scale heterogeneous graph datasets.}
\label{exp:heterogeneous oag}
\end{table*}

% Performances of Node Classification  (Heterogeneous)
\begin{table}[]
\centering
\begin{tabular}{ l c c c }
\toprule
\multicolumn{1}{ c }{Model} & DBLP & ACM  & IMDB \\ \midrule
MLP                          & 79.18$\pm$0.015 & 86.19$\pm$0.003 & 49.51$\pm$0.019 \\
Node2Vec                          & 86.10$\pm$0.001 & 76.27$\pm$0.002 & 47.32$\pm$0.005 \\
\midrule
GCN  & 85.63$\pm$0.003 & 91.70$\pm$0.003 & 60.41$\pm$0.009 \\
GAT  & 94.68$\pm$0.002 & 91.99$\pm$0.003 & 59.64$\pm$0.016 \\
RGCN  & 93.16$\pm$0.002 & 91.93$\pm$0.004 & 59.87$\pm$0.008 \\
HAN  & 92.17$\pm$0.005 & 91.10$\pm$0.004 & 59.80$\pm$0.013 \\ 
HGT  & 94.21$\pm$0.005 & 91.14$\pm$0.005 & 60.98$\pm$0.002 \\
\midrule
$\text{GTN}_{-I}$  & 94.74$\pm$0.005 & 85.36$\pm$0.021  & 59.27$\pm$0.021  \\
GTN  & 94.47$\pm$0.003 & 91.96$\pm$0.005 & 61.02$\pm$0.018 \\
FastGTN   & \textbf{94.85$\pm$0.003} & \textbf{92.51$\pm$0.005} & \textbf{64.63$\pm$0.008} \\
\bottomrule
\end{tabular}
\vspace{1mm}
\caption{Node classification (micro F1-score) on heterogeneous graph datasets.}
\vspace{-3mm}
\label{exp:heterogeneous}
\end{table}

\begin{table*}[t]
\centering
\begin{tabular}{ l c c c c c c }
\toprule
\multicolumn{1}{ c }{Model} & \textsc{Air-USA}        & \textsc{Blogc}          & \textsc{Citeseer}       & \textsc{Cora}           & \textsc{Flickr}         & \textsc{PPI} \\ \midrule
MLP                          & 55.10$\pm$0.006 & 82.42$\pm$0.010 & 59.23$\pm$0.007 & 55.78$\pm$0.014 & 69.10$\pm$0.009 & 40.38$\pm$0.001    \\
Node2Vec                          & 45.86$\pm$0.013 & 61.23$\pm$0.001 & 33.51$\pm$0.007 & 54.30$\pm$0.006 & 46.50$\pm$0.002 & 40.87$\pm$0.001    \\
\midrule
GCN       & 57.30$\pm$0.009 & 75.25$\pm$0.006 & 68.42$\pm$0.006 & 79.65$\pm$0.005 & 52.95$\pm$0.009 & 42.56$\pm$0.003    \\
GAT     & 53.06$\pm$0.001 & 55.63$\pm$0.021 & 68.92$\pm$0.006 & 79.85$\pm$0.009 & 36.10$\pm$0.017 & 40.63$\pm$0.013    \\ 
JK-NET         & 57.15$\pm$0.009 & 68.47$\pm$0.004 & 68.11$\pm$0.007 & 79.57$\pm$0.006 & 54.01$\pm$0.006 &    42.38$\pm$0.003 \\
MixHop       & 55.15$\pm$0.011 & 66.96$\pm$0.003 & 67.74$\pm$0.013 & 79.32$\pm$0.007 & 48.91$\pm$0.014 &    42.52$\pm$0.002 \\
GCNII      & 56.25$\pm$0.010 & 64.17$\pm$0.004 & 68.11$\pm$0.017 & 79.24$\pm$0.014 & 33.71$\pm$0.008 &    42.36$\pm$0.004 \\
\midrule
$\text{GTN}_{-I}$  & 61.51 $\pm$0.013 
                    & 60.73 $\pm$0.009               
                    & 64.64 $\pm$0.009 
                    & 76.97 $\pm$0.005 
                    & 30.65 $\pm$0.008 
                    & \textbf{42.95 $\pm$0.004}     \\
GTN                 & \textbf{61.80 $\pm$0.008}
                    & \textbf{90.30 $\pm$0.006}
                    & 68.68 $\pm$0.011 
                    & 79.99 $\pm$0.008
                    & \textbf{76.77 $\pm$0.009}
                    & 42.64 $\pm$0.003    \\
FastGTN             & 57.73$\pm$0.008               
                    & 87.97$\pm$0.008
                    & \textbf{69.14$\pm$0.016}
                    & \textbf{80.29$\pm$0.009}
                    & 73.64$\pm$0.015
                    & 42.40$\pm$0.010 \\
\bottomrule
\end{tabular}
\vspace{1mm}
\caption{Node classification performance (micro F1-score) on homogeneous graph datasets.}
\label{exp:homogeneous}
\end{table*}

\subsection{Results on Node Classification}
We evaluated the effectiveness of our GTNs and FastGTNs with non-local operations in six heterogeneous graph datasets and six homogeneous graph datasets. By analysing the result of our experiment, we will answer the research \textbf{Q1}.

\noindent \textbf{Effectiveness of Graph Transformer Networks on heterogeneous graph datasets.} \hjk{Table \ref{exp:heterogeneous oag}. and \ref{exp:heterogeneous}. show the classification results on six heterogeneous graph datasets. In large-scale graph datasets (e.g., CS, ML, NN, DBLP, BLOGCATLOG, and FLICKR), we trained GNN-based methods and GTN in the mini-batch setting with graph sampling algorithm \cite{hu2020hgt, hamilton2017inductive}.} \tpami{We observe that our propsed methods, GTN and FastGTN, consistently outperform all network embedding methods and graph neural network methods in six heterogeneous graph datasets. GNN-based methods perform better than random walk-based network embedding methods. Interestingly, though the HAN is a modified GAT for a heterogeneous graph, the GAT usually performs better than the HAN. This result shows that using the pre-defined meta-paths as the HAN may cause adverse effects on performance. In contrast, Our GTN and FastGTN achieved the best performance compared to all other baselines on all the datasets. 
It demonstrates that the GTN can learn a new graph structure which consists of useful meta-paths for learning more effective node representation. 
Also, the performance gap between GTNs and FastGTNs on IMDB (60.02\% vs. 64.64\%)
shows that in FastGTNs the graph transformations based on the semantic similarity (i.e., non-local operations) are effective. We additionally provide an ablation study of non-local operations in A.6 in the supplement.}

\noindent \textbf{Effectiveness of Graph Transformer Networks on homogeneous graphs.}
\hjk{In homogeneous graphs, although a number of edge types is only one, as we add an identity matrix to the candidate adjacency matrix, our GTNs can find the effective neighborhood range for each dataset.}
\tpami{Table \ref{exp:homogeneous} shows the performance of GTNs, FastGTNs and other baselines in homogeneous graphs. We additionally compared our methods with three well-known GNN models MixHop \cite{abu2019mixhop}, JK-Net \cite{xu2018jknet} and GCNII.  \cite{chen2020simple}. We can observe that our GTN and FastGTN consistently outperform all GNN baselines in homogeneous graph datasets, especially on \textsc{Blogcatalog} and \textsc{Flickr} datasets. Interestingly, in \textsc{Blogcatalog} and \textsc{Flickr} datasets, the Multi-Layer Perceptron (MLP) model using the \textit{only} node features achieved better performance than all GNN baseline models.
It implies that noisy input graphs rather hinder learning of most GNNs whereas GTNs and FastGTNs successfully suppress the noisy edges by weighing an attention score of an identity matrix and learn a high accuracy classifier. }

\noindent \textbf{Identity matrix in $\mathbb{A}$ to learn variable-length meta-paths}. As mentioned in Section \ref{sub:Learning Meta-Path Graphs}, the identity matrix is included in the candidate adjacency matrices $\mathbb{A}$. 
To verify the effect of identity matrix, we trained and evaluated another model named  $GTN_{-I}$ as an ablation study.  \ysj{the} $GTN_{-I}$ has exactly the same model structure as GTN but its candidate adjacency matrix $\mathbb{A}$ doesn't include an identity matrix. In general, \ysj{the} $GTN_{-I}$ usually performs worse than \ysj{the} GTN.  In heterogeneous graph datasets, it is worth to note that the difference is greater in IMDB than DBLP.
One explanation is that the length of meta-paths $GTN_{-I}$ produced is not effective in IMDB. As we stacked 3 layers of GTL, $GTN_{-I}$ always produce 4-length meta-paths. However shorter meta-paths (e.g. MDM) are preferable in IMDB. 
\tpami{Also, in homogeneous graph datasets, the differences in \textsc{BlogCatalog} and \textsc{Flickr} are extremely big, which are $48\%$ and $150\%$. It demonstrates that including identity matrix is effective in learning GTNs on noisy graphs.}

% Identity_Figure
\begin{figure}[h]
 \centering
 \includegraphics[width=150px]{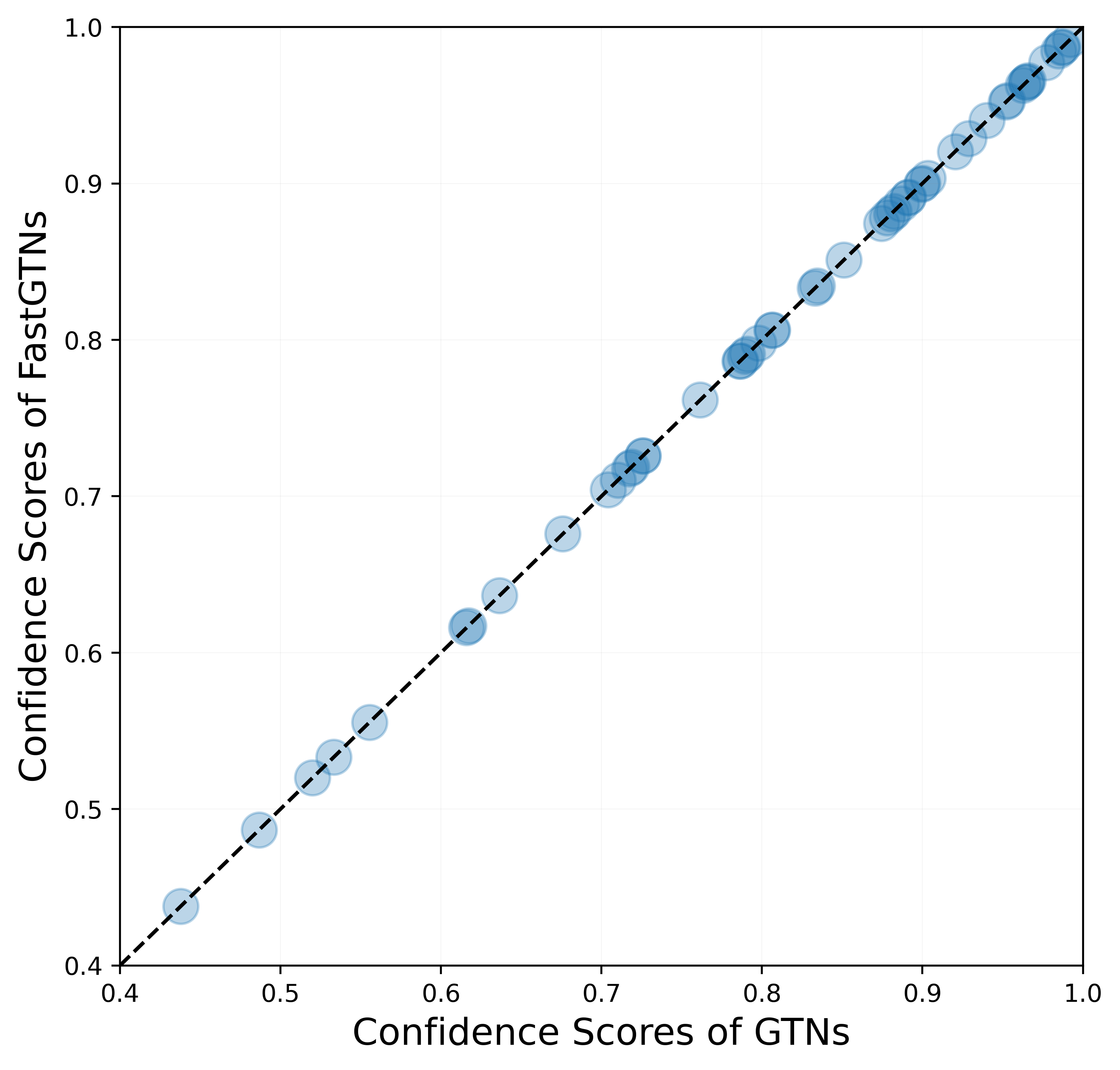}
 \caption{Confidence scores of GTNs (x-axis) and FastGTNs (y-axis). On 50 random samples from the test set of IMDB, the confidence scores of GTNs and FastGTNs are practically identical. All 50 points are on the Identity line i.e., $y=x$.}
 \label{fig:gtn_confidence scores}
\end{figure}

\begin{figure*}[t]
 \centering
 \includegraphics[width=350px]{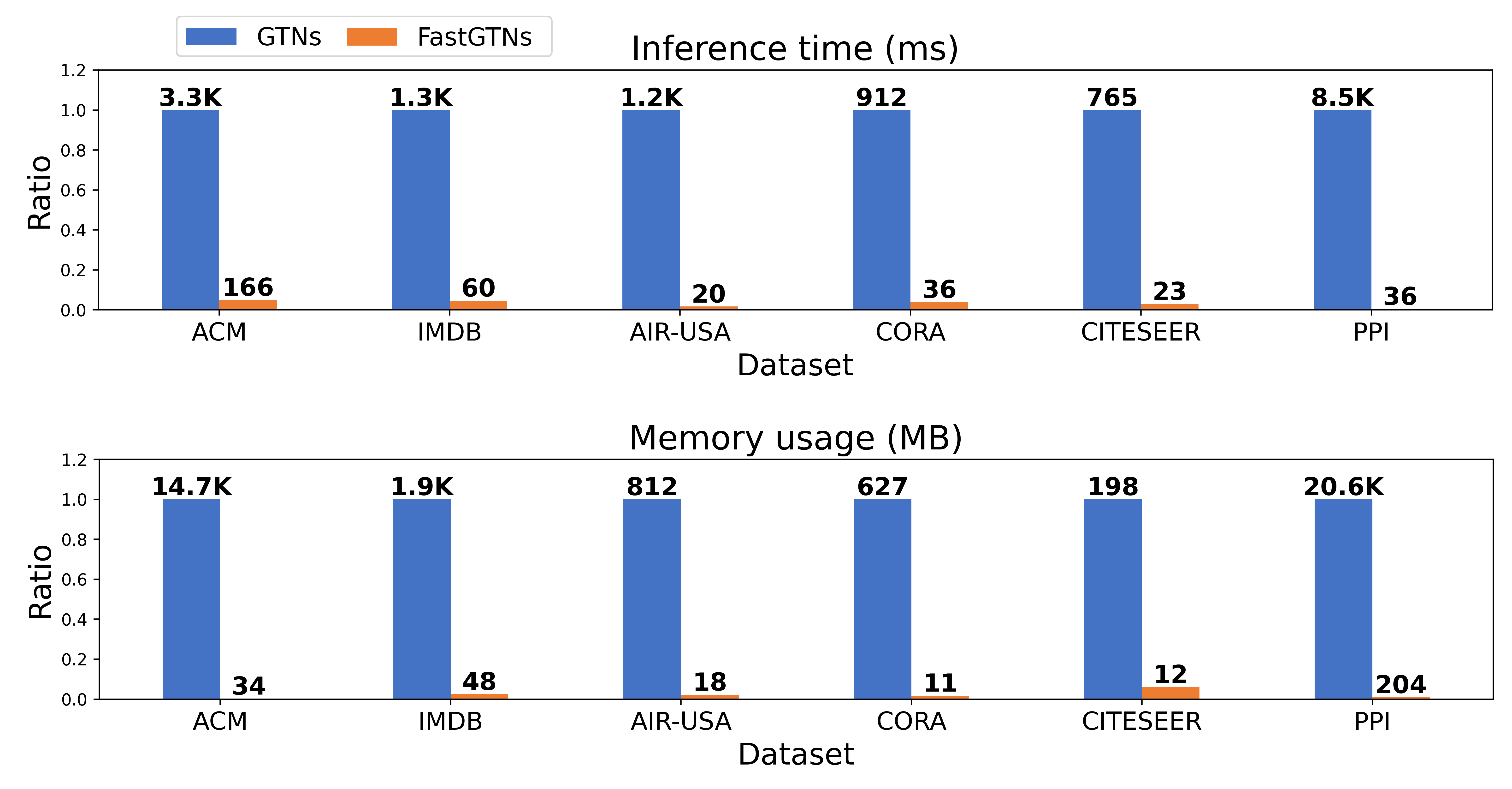}
 \caption{Inference time (up) and memory usage (down) of GTNs (blue) and our FastGTNs (orange) on both homogeneous and heterogeneous graph datasets.
 To show how efficiently FastGTNs can perform the \textit{identical} graph transformations, FastGTNs were measured without the non-local operations. 
 FastGTNs significantly speed up the graph transformations and reduce the memory usuage in all datasets. Especially, on a large-scale graph dataset (PPI) FastGTNs show 230$\times$ faster inference time and 100$\times$ less memory usuage than the GTNs.}
 \label{fig:scalability}
\end{figure*}

\begin{table*}[t]\label{tab:mp_anl}
\centering
\begin{tabular}{cccccccc}
\toprule
\multirow{2}{*}{Dataset}% & \multicolumn{3}{c}{Meta-Paths} \\[0.2em]%\cmidrule{3-9}
& Predefined &  \multicolumn{2}{c}{Meta-path learnt by GTNs}\\\
& Meta-path & Top 3 (between target nodes) & Top 3 (all)\\\midrule
\multirow{1}{*}{DBLP} & APCPA, APA  & APCPA, APAPA, APA & CPCPA, APCPA, CP \\
\\[-0.5em]
\multirow{1}{*}{ACM} & PAP, PSP & PAP, PSP & APAP, APA, SPAP \\
\\[-0.5em]
\multirow{1}{*}{IMDB} & MAM, MDM & MDM, MAM, MDMDM & DM, AM, MDM\\\bottomrule
\end{tabular}
\vspace{1mm}
\caption{Comparison with predefined meta-paths and top-ranked meta-paths by GTNs. Our model found important meta-paths that are consistent with pre-defined meta-paths between target nodes (a type of nodes with labels for node classifications). Also, new relevant meta-paths between all types of nodes are discovered by GTNs.}
\label{table:evaluation_results}
\end{table*}

\medskip

\subsection{Exactness and Efficiency of FastGTNs}
\label{sub:identity btw GTN and FastGTN}
\tpami{In this section, we show the exactness and efficiency of FastGTN compared to GTN.
First, As discussed in \eqref{eq:associative} in Section \ref{sec:3.2}, Our FastGTN without non-local operations is an exact version of GTN. 
It means that the model parameters from GTN are compatible with FastGTN
and the graph transformations of GTN can be \textit{identically} performed by FastGTN.
To show the exactness of FastGTN, we first train a GTN on the IMDB dataset and 
copy the model parameters of the GTN ($\{\Phi^{(k)}\}_{k=1}^K$) and the GNN $\{W^{(l)}\}_{l=1}^L$ to the corresponding model parameters in our FastGTN.
Figure \ref{fig:gtn_confidence scores} proves that the predictions (confidence scores) by the FastGTN and the GTN are identical.
All 50 randomly drawn data points from a test set are on the Identity line (i.e., $y=x$).
Note that the GTN parameters $\{\Phi^{(k)}\}_{k=1}^K$ should be reversely copied as in \eqref{eq:FastGTN}.
As our FastGTNs also generalize of other popular graph neural networks such as GCN and MixHop, we prove that predictions (confidence scores) by special cases of our FastGTNs and the two GNNs are identical (see A.5 in the supplement).} %~\ref{fig:gcn_confidence scores} in the supplement).

\tpami{Second, the main contribution of our FastGTN is improving the efficiency of graph transformations. 
For more detailed efficiency comparisons with GTN, we measured the inference time and memory consumption of the two methods. 
Figure~\ref{fig:scalability} shows that FastGTN is significantly more efficient than GTN in both the inference time and memory consumption. The performance gain is larger on larger graphs.
In particular, on the \textsc{PPI} dataset, our experiments show that our FastGTN is 230$\times$ faster and 100$\times$ more memory-efficient than the GTN. Again, this speed-up and memory efficiency are achieved without any accuracy loss.
During training, a similar performance gain is observed (see A.4 in the supplement).}

\begin{figure*}[!t]
    \centering
    \subfloat[DBLP]{\includegraphics[trim={30, 20, 30, 20}, clip=True, width=180pt, height=180pt]{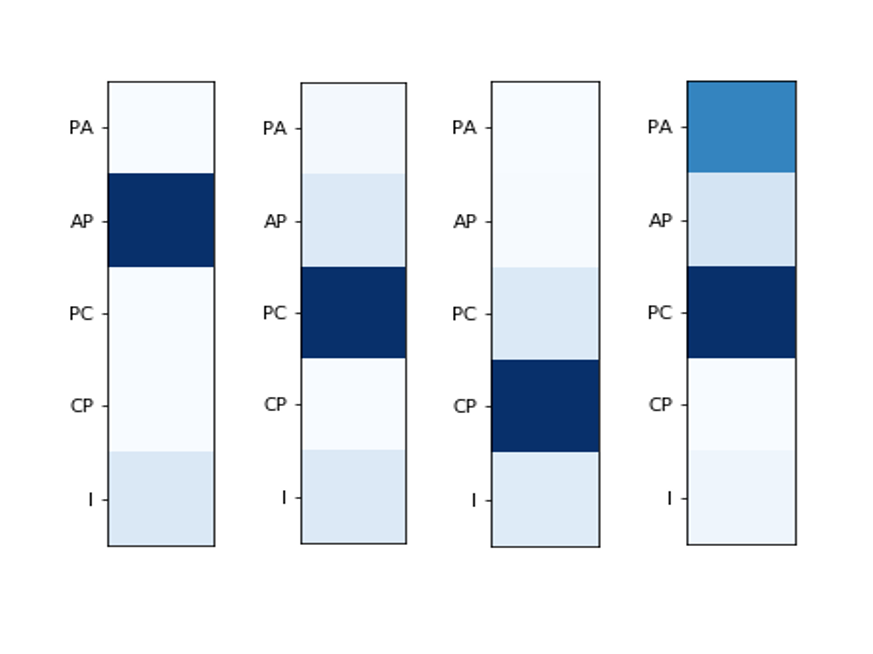}}
    \label{fig:nrGroup}
    \hspace{1.5cm}
    \centering
    % \hspace{1.5cm}
    \subfloat[IMDB]{\includegraphics[trim={30, 20, 30, 20}, clip=True, width=180pt, height=180pt]{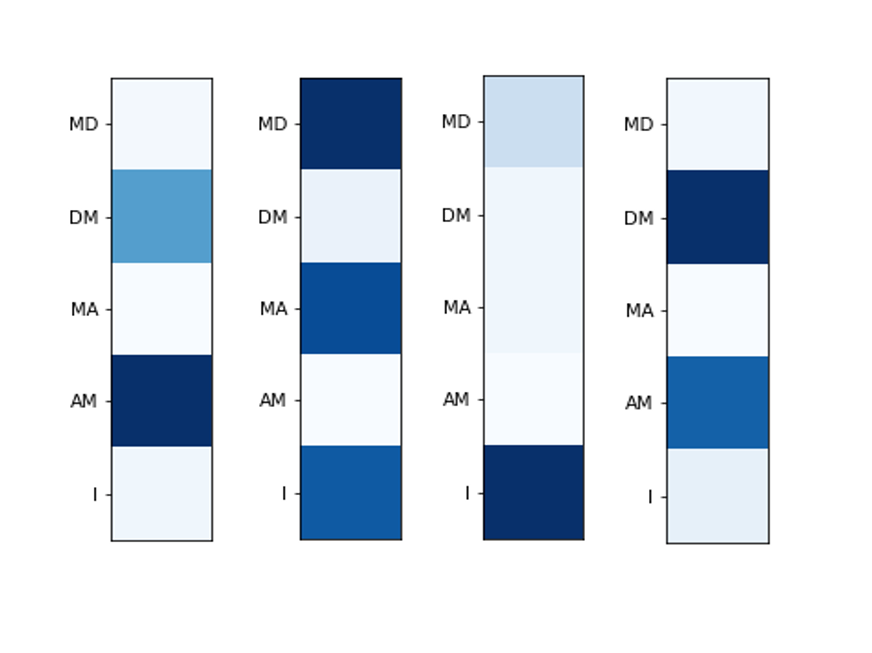}}
    \label{fig:overallResult}
    \caption{
    After applying softmax function on 1x1 conv filter $phi^{(k)}$ (k: index of layer) in Figure \ref{fig:GTLayer}, we visualized this attention score of adjacency matrix (edge type) in DBLP (left) and IMDB (right) datasets. (a) Respectively, each edge indicates (Paper-Author), (Author-Paper), (Paper-Conference), (Conference-Paper), and identity matrix. (b) Edges in IMDB dataset indicates (Movie-Director), (Director-Movie), (Movie-Actor), (Actor-Movie), and identity matrix.
    }
    \label{fig:attention_score}
\end{figure*}

%\subsection{Analysis of Learning the structure of the graph}
\subsection{Interpretation of Graph Transformer Networks}
We examine the transformation learnt by GTNs to discuss the question interpretability $\textbf{Q4}$. 
We first describe how to calculate the importance of each meta-path from our GT layers. 
For the simplicity, we assume the number of output channels is one.
Then, the new adjacency matrix from the $l$th GT layer can be written as 
\begin{align}
    \label{eq:metp_att}
    \begin{split}
        {A}^{(k)} &= \invD{k-1}\ldots{\invD{1}\bigg( (\alpha^{(0)}\cdot \mathbb{A}) (\alpha^{(1)}\cdot \mathbb{A})}   \\
        &\phantom{{}={}} \ldots (\alpha^{(k)}\cdot \mathbb{A}) \bigg) \\
        % &\phantom{{}={}} (\alpha^{(2)} \cdot \mathbb{A}) \ldots (\alpha^{(l)}\cdot \mathbb{A}) \bigg)} \\
        &= \invD{k-1}\ldots{\invD{1}\bigg(\sum_{t_0,t_1,...,t_k \in \mathcal{T}^e}{\alpha_{t_0}^{(0)} \alpha_{t_1}^{(1)}}} \\
        &\phantom{{}={}} \dots \alpha_{t_k}^{(k)}A_{t_0}A_{t_{1}}\dots A_{t_k}\bigg),
    \end{split}
\end{align}

\noindent where $\mathcal{T}^e$ denotes a set of edge types and $\alpha_{t_l}^{(l)}$ is an attention score for edge type $t_l$ at the $l$th GT layer. So, $A^{(l)}$ can be viewed as a weighted sum of all meta-paths including $1$-length (original edges) to $l$-length meta-paths. The contribution of a meta-path $t_l,t_{l-1}, \ldots ,t_0$ is obtained by $\prod_{i=0}^{l} \alpha_{t_i}^{(i)}$.

 Now we can interpret new graph structures learnt by GTNs. The weight $\prod_{i=0}^{l} \alpha_{t_i}^{(i)}$ for a meta-path ($t_0, t_1, \ldots t_l$) is an attention score and it provides the importance of the meta-path in the prediction task.
 In Table \ref{table:evaluation_results} we summarized predefined meta-paths, that are widely used in literature, and the meta-paths with high attention scores learnt by GTNs. 

As shown in Table \ref{table:evaluation_results}, between target nodes, that have class labels to predict, the predefined meta-paths by domain knowledge are consistently top-ranked by GTNs as well. 
This shows that GTNs are capable of learning the importance of meta-paths for tasks. 
More interestingly, GTNs discovered important meta-paths that are not in the predefined meta-path set.
For example, in the DBLP dataset GTN ranks CPCPA as most importance meta-paths, which is not included in the predefined meta-path set. It makes sense that author's research area (label to predict) is relevant to the venues where the author publishes.
We believe that the interpretability of GTNs provides useful insight in node classification by the attention scores on meta-paths.  

%\subsection{Visualization of Meta-Paths and Adjacency Matrix}
Fig.~\ref{fig:attention_score} shows the attention scores of adjacency matrices (edge type) from each Graph Transformer Layer. Compared to the result of DBLP, identity matrices have higher attention scores in IMDB. 
As discussed in Section \ref{method-graph_transformer}, a GTN is capable of learning shorter meta-paths than the number of GT layers\sjy{, which they} are more effective as in IMDB. 
By assigning higher attention scores to the identity matrix, the GTN tries to stick to the shorter meta-paths even in the deeper layer. This result demonstrates that the GTN has ability to adaptively learns most effective meta-path length depending on the dataset. Also, we additionally provide an interpretation of GTNs in homogeneous graphs in A.7 in the supplement.

\section{Conclusion}
We proposed Graph Transformer Networks for learning node representations on both homogeneous graphs and heterogeneous graphs.
Our approach transforms graphs into multiple new graphs defined by meta-paths with arbitrary edge types and arbitrary length up to one less than the number of Graph Transformer layers while it learns node representation via convolution on the learnt meta-path graphs. 
Also, we proposed the enhanced version of GTNs, Fast Graph Transformer Networks, which are 230× faster and use 100× less memory while allowing the identical graph transformations as GTNs.
The learnt graph structures from GTNs and FastGTNs lead to more effective node representation resulting in state-of-the art performance, without any predefined meta-paths from domain knowledge, on all twelve benchmark node classification on both homogeneous and heterogeneous graphs.

Since our Graph Transformer layers can be combined with existing GNNs, we believe that our framework opens up new ways for GNNs to optimize graph structures by themselves to operate convolution depending on data and tasks without any manual efforts.
As several heterogeneous graph datasets have been recently studied for other network analysis tasks, such as link prediction \cite{kgat_kdd19,hetgnn_kdd19} and graph classification \cite{kim2019hats, hgat_emnlp19}, applying our GTNs to the other tasks can be interesting future directions.

\ifCLASSOPTIONcaptionsoff
  \newpage
\fi

\bibliographystyle{unsrt}
\bibliography{ms}

\begin{thebibliography}{10}

\bibitem{kipf2016gcn}
Thomas~N Kipf and Max Welling.
\newblock Semi-supervised classification with graph convolutional networks.
\newblock {\em International Conference on Learning Representations}, 2017.

\bibitem{li2018deeper}
Qimai Li, Zhichao Han, and Xiao ming Wu.
\newblock Deeper insights into graph convolutional networks for semi-supervised
  learning, 2018.

\bibitem{xu2018jknet}
Keyulu Xu, Chengtao Li, Yonglong Tian, Tomohiro Sonobe, Ken-ichi Kawarabayashi,
  and Stefanie Jegelka.
\newblock Representation learning on graphs with jumping knowledge networks.
\newblock {\em International Conference on Machine Learning}, 2018.

\bibitem{liu2019geniepath}
Ziqi Liu, Chaochao Chen, Longfei Li, Jun Zhou, Xiaolong Li, Le~Song, and Yuan
  Qi.
\newblock Geniepath: Graph neural networks with adaptive receptive paths.
\newblock In {\em Proceedings of the AAAI Conference on Artificial
  Intelligence}, 2019.

\bibitem{liu2020deepergnn}
Meng Liu, Hongyang Gao, and Shuiwang Ji.
\newblock Towards deeper graph neural networks.
\newblock In {\em Proceedings of the 26th ACM SIGKDD International Conference
  on Knowledge Discovery \& Data Mining}, 2020.

\bibitem{kipf2016variational}
Thomas~N Kipf and Max Welling.
\newblock Variational graph auto-encoders.
\newblock {\em NIPS Workshop on Bayesian Deep Learning}, 2016.

\bibitem{schlichtkrull2018rgcn}
Michael Schlichtkrull, Thomas~N Kipf, Peter Bloem, Rianne Van Den~Berg, Ivan
  Titov, and Max Welling.
\newblock Modeling relational data with graph convolutional networks.
\newblock In {\em European Semantic Web Conference}. Springer, 2018.

\bibitem{zhang2018link}
Muhan Zhang and Yixin Chen.
\newblock Link prediction based on graph neural networks.
\newblock In {\em Advances in Neural Information Processing Systems}, 2018.

\bibitem{teru2019inductive}
Komal~K Teru, Etienne Denis, and William~L Hamilton.
\newblock Inductive relation prediction by subgraph reasoning.
\newblock In {\em Proceedings of the International Conference on Machine
  Learning}, 2020.

\bibitem{vashishth2019compgcn}
Shikhar Vashishth, Soumya Sanyal, Vikram Nitin, and Partha Talukdar.
\newblock Composition-based multi-relational graph convolutional networks.
\newblock In {\em International Conference on Learning Representations}, 2020.

\bibitem{ying2018diffpool}
Zhitao Ying, Jiaxuan You, Christopher Morris, Xiang Ren, Will Hamilton, and
  Jure Leskovec.
\newblock Hierarchical graph representation learning with differentiable
  pooling.
\newblock In {\em Advances in neural information processing systems}, 2018.

\bibitem{xu2018powerful}
Keyulu Xu, Weihua Hu, Jure Leskovec, and Stefanie Jegelka.
\newblock How powerful are graph neural networks?
\newblock In {\em International Conference on Learning Representations}, 2019.

\bibitem{gao2019graphunet}
Hongyang Gao and Shuiwang Ji.
\newblock Graph u-nets.
\newblock In {\em International Conference on Machine Learning}, 2019.

\bibitem{ma2019graph}
Yao Ma, Suhang Wang, Charu~C Aggarwal, and Jiliang Tang.
\newblock Graph convolutional networks with eigenpooling.
\newblock In {\em Proceedings of the 25th ACM SIGKDD International Conference
  on Knowledge Discovery \& Data Mining}, 2019.

\bibitem{you2018graphrnn}
Jiaxuan You, Rex Ying, Xiang Ren, William~L Hamilton, and Jure Leskovec.
\newblock Graphrnn: Generating realistic graphs with deep auto-regressive
  models.
\newblock In {\em International Conference on Machine Learning}, 2018.

\bibitem{li2018generative}
Yujia Li, Oriol Vinyals, Chris Dyer, Razvan Pascanu, and Peter Battaglia.
\newblock Learning deep generative models of graphs.
\newblock In {\em International Conference on Machine Learning}, 2018.

\bibitem{you2018graph}
Jiaxuan You, Bowen Liu, Zhitao Ying, Vijay Pande, and Jure Leskovec.
\newblock Graph convolutional policy network for goal-directed molecular graph
  generation.
\newblock In {\em Advances in neural information processing systems}, 2018.

\bibitem{liao2019efficient}
Renjie Liao, Yujia Li, Yang Song, Shenlong Wang, Will Hamilton, David~K
  Duvenaud, Raquel Urtasun, and Richard Zemel.
\newblock Efficient graph generation with graph recurrent attention networks.
\newblock In {\em Advances in Neural Information Processing Systems}, 2019.

\bibitem{dai2020scalable}
Hanjun Dai, Azade Nazi, Yujia Li, Bo~Dai, and Dale Schuurmans.
\newblock Scalable deep generative modeling for sparse graphs.
\newblock {\em International Conference on Machine Learning}, 2020.

\bibitem{wang2019han}
Xiao Wang, Houye Ji, Chuan Shi, Bai Wang, Yanfang Ye, Peng Cui, and Philip~S
  Yu.
\newblock Heterogeneous graph attention network.
\newblock In {\em The World Wide Web Conference}, 2019.

\bibitem{zhang2018deep}
Yizhou Zhang, Yun Xiong, Xiangnan Kong, Shanshan Li, Jinhong Mi, and Yangyong
  Zhu.
\newblock Deep collective classification in heterogeneous information networks.
\newblock In {\em Proceedings of the 2018 World Wide Web Conference on World
  Wide Web}, 2018.

\bibitem{bruna2013spectral}
Joan Bruna, Wojciech Zaremba, Arthur Szlam, and Yann Lecun.
\newblock Spectral networks and locally connected networks on graphs.
\newblock In {\em International Conference on Learning Representations}, 2014.

\bibitem{defferrard2016convolutional}
Micha{\"e}l Defferrard, Xavier Bresson, and Pierre Vandergheynst.
\newblock Convolutional neural networks on graphs with fast localized spectral
  filtering.
\newblock In {\em Advances in neural information processing systems}, 2016.

\bibitem{hamilton2017inductive}
Will Hamilton, Zhitao Ying, and Jure Leskovec.
\newblock Inductive representation learning on large graphs.
\newblock In {\em Advances in neural information processing systems}, 2017.

\bibitem{velivckovic2017gat}
Petar Veli{\v{c}}kovi{\'c}, Guillem Cucurull, Arantxa Casanova, Adriana Romero,
  Pietro Lio, and Yoshua Bengio.
\newblock Graph attention networks.
\newblock {\em International Conference on Learning Representations}, 2018.

\bibitem{abu2019mixhop}
Sami Abu-El-Haija, Bryan Perozzi, Amol Kapoor, Nazanin Alipourfard, Kristina
  Lerman, Hrayr Harutyunyan, Greg~Ver Steeg, and Aram Galstyan.
\newblock Mixhop: Higher-order graph convolutional architectures via sparsified
  neighborhood mixing.
\newblock {\em International Conference on Machine Learning}, 2019.

\bibitem{hu2020hgt}
Ziniu Hu, Yuxiao Dong, Kuansan Wang, and Yizhou Sun.
\newblock Heterogeneous graph transformer.
\newblock In {\em Proceedings of The Web Conference 2020}, 2020.

\bibitem{vaswani2017attention}
Ashish Vaswani, Noam Shazeer, Niki Parmar, Jakob Uszkoreit, Llion Jones,
  Aidan~N Gomez, {\L}ukasz Kaiser, and Illia Polosukhin.
\newblock Attention is all you need.
\newblock In {\em International Conference on Machine Learning}, 2017.

\bibitem{shi2016survey}
Chuan Shi, Yitong Li, Jiawei Zhang, Yizhou Sun, and S~Yu Philip.
\newblock A survey of heterogeneous information network analysis.
\newblock {\em IEEE Transactions on Knowledge and Data Engineering}, 2016.

\bibitem{kipf2016semi}
Thomas~N Kipf and Max Welling.
\newblock Semi-supervised classification with graph convolutional networks.
\newblock {\em arXiv preprint arXiv:1609.02907}, 2016.

\bibitem{tang2015line}
Jian Tang, Meng Qu, Mingzhe Wang, Ming Zhang, Jun Yan, and Qiaozhu Mei.
\newblock Line: Large-scale information network embedding.
\newblock In {\em WWW}, 2015.

\bibitem{wu2019net}
Jun Wu, Jingrui He, and Jiejun Xu.
\newblock Net: Degree-specific graph neural networks for node and graph
  classification.
\newblock In {\em Proceedings of the 25th ACM SIGKDD International Conference
  on Knowledge Discovery \& Data Mining}, 2019.

\bibitem{huang2017label}
Xiao Huang, Jundong Li, and Xia Hu.
\newblock Label informed attributed network embedding.
\newblock In {\em Proceedings of the Tenth ACM International Conference on Web
  Search and Data Mining}, 2017.

\bibitem{grover2016node2vec}
Aditya Grover and Jure Leskovec.
\newblock node2vec: Scalable feature learning for networks.
\newblock In {\em Proceedings of the 22nd ACM SIGKDD international conference
  on Knowledge discovery and data mining}, 2016.

\bibitem{chen2020simple}
Ming Chen, Zhewei Wei, Zengfeng Huang, Bolin Ding, and Yaliang Li.
\newblock Simple and deep graph convolutional networks.
\newblock In {\em International Conference on Machine Learning}, 2020.

\bibitem{fey2019torchgeometric}
Adam Paszke, Sam Gross, Francisco Massa, Adam Lerer, James Bradbury, Gregory
  Chanan, Trevor Killeen, Zeming Lin, Natalia Gimelshein, Luca Antiga, Alban
  Desmaison, Andreas Kopf, Edward Yang, Zachary DeVito, Martin Raison, Alykhan
  Tejani, Sasank Chilamkurthy, Benoit Steiner, Lu~Fang, Junjie Bai, and Soumith
  Chintala.
\newblock Pytorch: An imperative style, high-performance deep learning library.
\newblock In {\em Advances in Neural Information Processing Systems}. 2019.

\bibitem{kgat_kdd19}
Xiang Wang, Xiangnan He, Yixin Cao, Meng Liu, and Tat-Seng Chua.
\newblock Kgat: Knowledge graph attention network for recommendation.
\newblock In {\em Proceedings of the 25th ACM SIGKDD International Conference
  on Knowledge Discovery \& Data Mining}, 2019.

\bibitem{hetgnn_kdd19}
Chuxu Zhang, Dongjin Song, Chao Huang, Ananthram Swami, and Nitesh~V Chawla.
\newblock Heterogeneous graph neural network.
\newblock In {\em Proceedings of the 25th ACM SIGKDD International Conference
  on Knowledge Discovery \& Data Mining}, 2019.

\bibitem{kim2019hats}
Raehyun Kim, Chan~Ho So, Minbyul Jeong, Sanghoon Lee, Jinkyu Kim, and Jaewoo
  Kang.
\newblock Hats: A hierarchical graph attention network for stock movement
  prediction, 2019.

\bibitem{hgat_emnlp19}
Hu~Linmei, Tianchi Yang, Chuan Shi, Houye Ji, and Xiaoli Li.
\newblock Heterogeneous graph attention networks for semi-supervised short text
  classification.
\newblock In {\em Proceedings of the 2019 Conference on Empirical Methods in
  Natural Language Processing}, 2019.

\end{thebibliography}

% \bibliography{ref}


% \end{thebibliography}

% \begin{IEEEbiography}[{\includegraphics[width=1in,height=1.25in,clip,keepaspectratio]{mshell}}]{Michael Shell}
\begin{IEEEbiography}[{\includegraphics[width=1in,height=1.25in,clip,keepaspectratio]{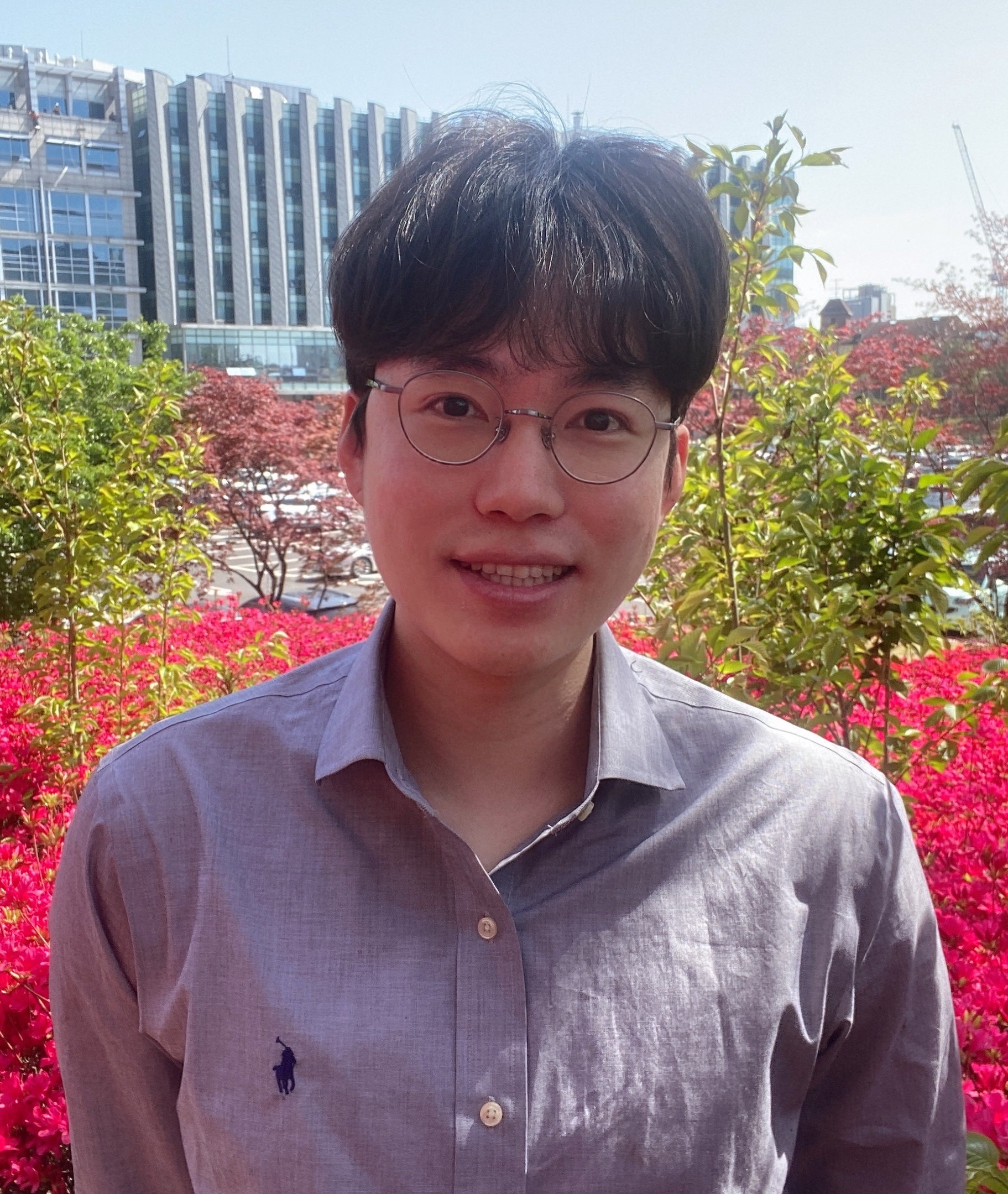}}]{Seongjun Yun}
received the bachelor’s degree from Computer Science and Engineering Department, Korea University, South Korea, in 2018.
He is working toward the doctoral degree at Korea University.
His current research interest includes graph neural networks, recommendation systems, and machine learning.
He was a recipient of the Naver PhD Fellowship award in 2020.
\end{IEEEbiography}

\begin{IEEEbiography}[{\includegraphics[width=1in,height=1.25in,clip,keepaspectratio]{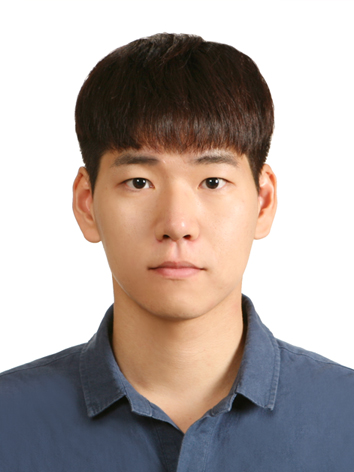}}]{Minbyul Jeong}
received the BS degree in Computer Science and Engineering Department from Korea University, Seoul, Republic of Korea. His current research interests include natural language processing for real world problems and graph neural network.
\end{IEEEbiography}

\begin{IEEEbiography}[{\includegraphics[width=1in,height=1.25in,clip,keepaspectratio]{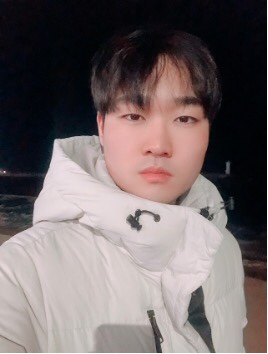}}]{Sungdong Yoo}
received the BS degree in Computer Science and Engineering Department, Korea University, in 2019. He is working toward the Master degree with Korea Univ. His research interest includes graph neural network and reinforcement learning. He was a research intern with LG CNS, Seoul, in 2019.
\end{IEEEbiography}

\begin{IEEEbiography}[{\includegraphics[width=1in,height=1.25in,clip,keepaspectratio]{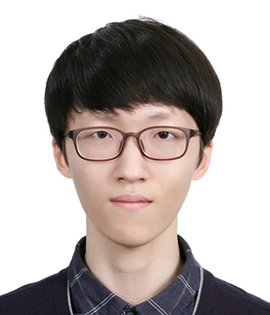}}]{Seunghun Lee}
received the BS degree from the Korea University in 2020, and he is currently working toward the MS degree in the Department of Computer Science and Engineering. His research interests are in neural networks, especially in graph neural networks and semi-supervised learning.
\end{IEEEbiography}

\begin{IEEEbiography}[{\includegraphics[width=1in,height=1.25in,clip,keepaspectratio]{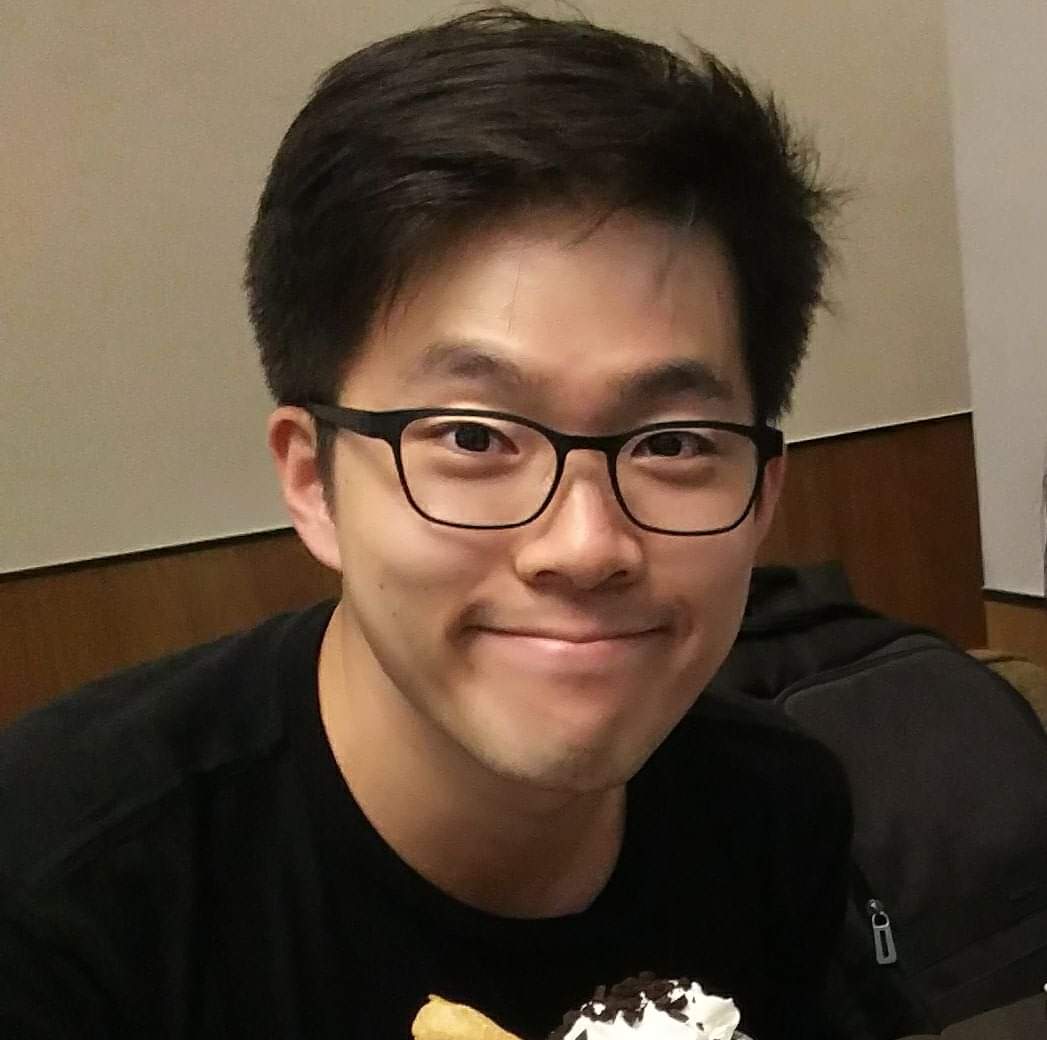}}]{Sean S. Yi}
received his bachelor's degree from Korea University in 2019, and is currently pursuing a master's degree from the same institution under the supervision of Professor Jaewoo Kang. His research interests largely lie in natural language processing, graphical models, and the interplay between the two. Specifically, he is interested in how to utilize graphical models in order to effectively model representations of human language and extract information.
\end{IEEEbiography}

\begin{IEEEbiography}[{\includegraphics[width=1in,height=1.25in,clip,keepaspectratio]{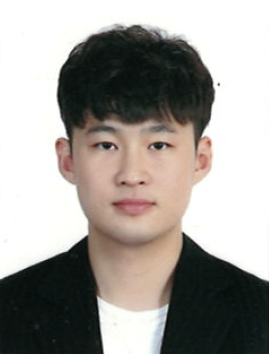}}]{Raehyun Kim}
received a B.S. degree in business from Korea University, Seoul, Republic of Korea,
in 2018, where he is currently pursuing an Ph.D. degree in computer science. 
His current research interests include financial market prediction and recommendation systems
\end{IEEEbiography}

%\newpage

\begin{IEEEbiography}[{\includegraphics[width=1in,height=1.25in,clip,keepaspectratio]{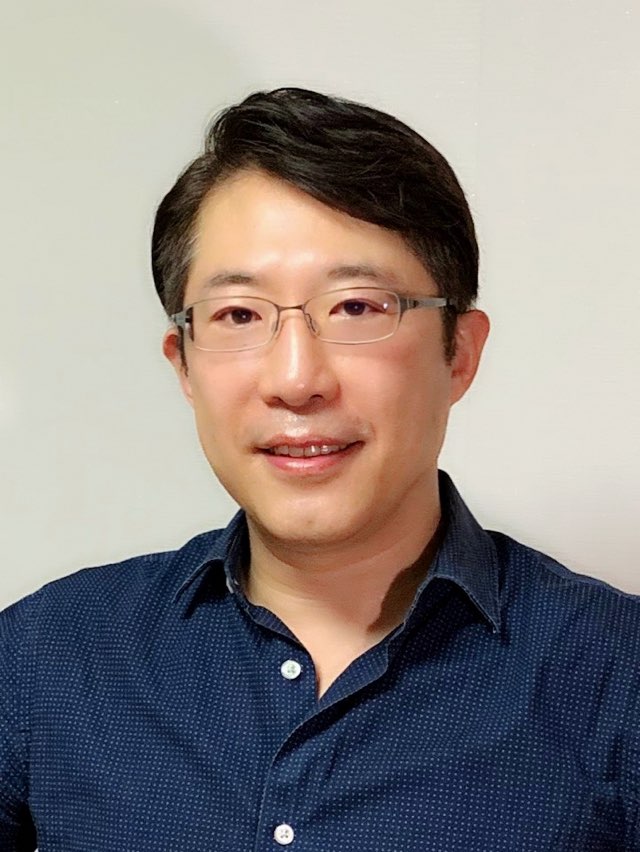}}]{Jaewoo Kang}
Jaewoo Kang (Member, IEEE) received B.S. degree in Computer Science from Korea University in 1994, M.S. in Computer Science from the University of Colorado at Boulder in 1996, and Ph.D. in Computer Science from the University of Wisconsin–Madison in 2003. From 1996 to 1997, he was a member of the technical staff at AT\&T Labs Research, Florham Park, NJ. From 2000 to 2001, he was CTO and Co-Founder of WISEngine Inc., Santa Clara, CA, and Seoul, Korea. From 2003 to 2006, he was an assistant professor in the computer science department at North Carolina State University. Since 2006, he has been a professor of Computer Science at Korea University. He is also head of the department of the interdisciplinary graduate program in Bioinformatics at Korea University. He is jointly appointed professor in the department of Medicine at Korea University.
\end{IEEEbiography}

\begin{IEEEbiography}[
    {\includegraphics[width=1in,height=1.1in,clip, trim=95 80 115 40,keepaspectratio]{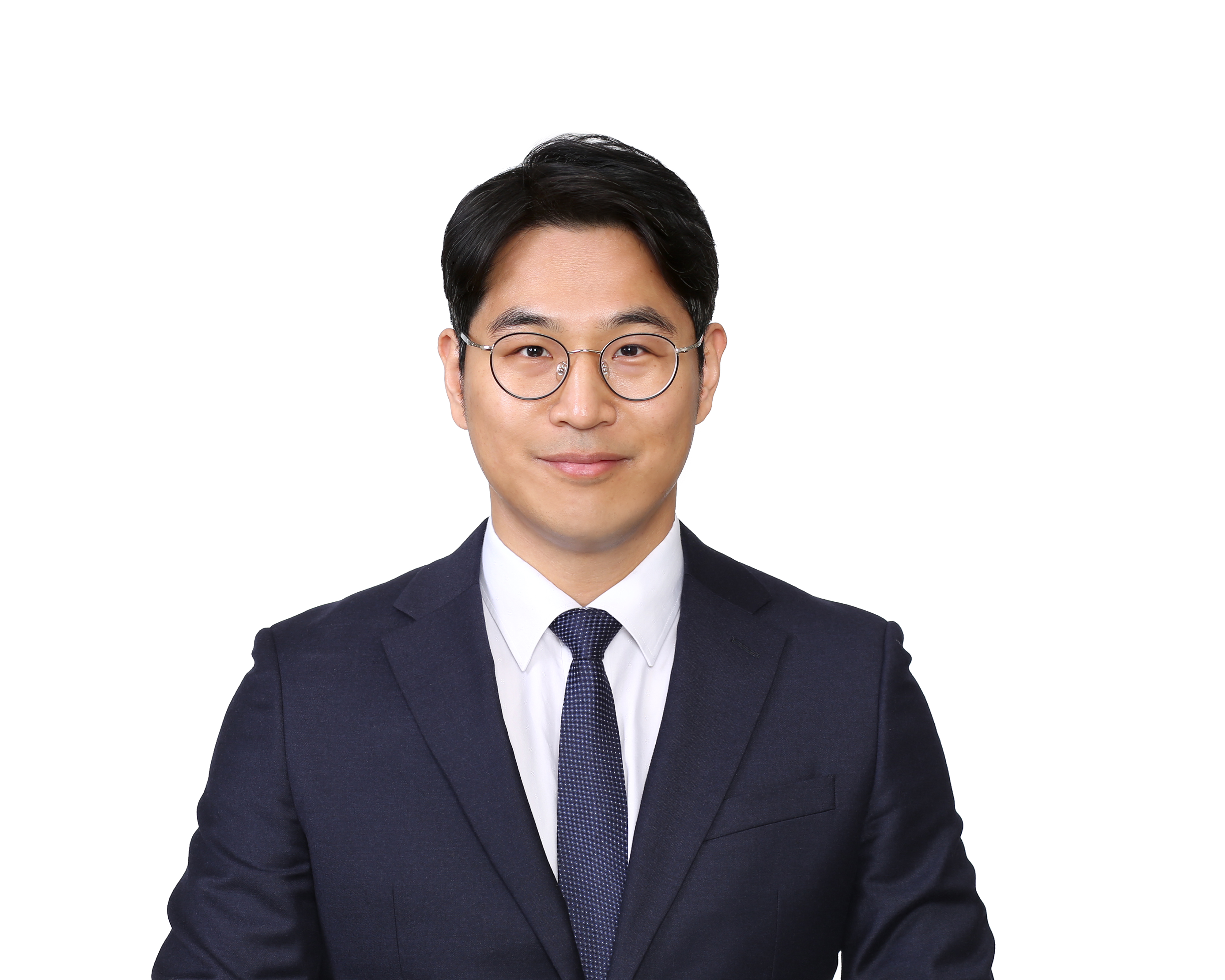}}]
{Hyunwoo J. Kim} is an assistant professor in the Department of Computer Science and Engineering at Korea University. Prior to the position, he worked at Amazon Lab126 in Sunnyvale California. He earned his Ph.D. in computer science in 2017 from the University of Wisconsin-Madison and his Ph.D. minor is statistics. His research interests include geometric deep learning for graphs and manifolds, manifold statistics, machine learning, computer vision, and medical imaging.
\end{IEEEbiography}
%%%%%%%%%%%%%%%%%%%%%%%%%%%%%%%%%%%%%%%%%%%%%%%%%%%%%%%


%\vfill


\newpage
\appendices
\section{}
\subsection{Mini-batch training for GTNs}
\label{app:Mini-batch training for GTNs}
Since GTNs transform the entire input graph \textit{at once}, when the size of an input graph is too large, GTNs requires an excessive amount of memory and incurs high computational cost. To alleviate this scalability issue, we present a mini-batch training algorithm for GTNs. Algorithm 1 describes our mini-batch training algorithm. Specifically, we first combine a set of input adjacency matrices $\{A_t\}_{t=1}^{T_e}$ into one adjacency matrix $A$ for extracting subgraphs. We then select target nodes for each mini-batch training iteration. For each iteration, based on the target nodes of each mini-batch, we extract a subgraph and use its adjacency matrix $A'$ from the original graph's adjacency matrix $A$ by using two types of graph sampling algorithms: a neighborhood-based graph sampling algorithm \cite{hamilton2017inductive} and a layer-wise graph sampling algorithm \cite{hu2020hgt}. After sampling the subgraph, we separate the sampled subgraph's adjacency matrix $A'$ into a set of adjacency matrices corresponding to each edge type such as $\{A'_t\}_{t=1}^{T_e}$. The sampled adjacency matrix set $\{A'_t\}_{t=1}^{T_e}$ is fed into GTNs for mini-batch training. This mini-batch training algorithm enables GTNs to handle large-scale graph datasets with up to 30 million edges in an efficient manner.
\begin{algorithm}
%\caption{Graph construction algorithm for minibatch training}
%\hspace*{\algorithmicindent} \textbf{Input}
%\hspace*{\algorithmicindent} \textbf{Output:} a set of adjacency matrices $\mathbb{A}$ of a heterogeneous graph G }
% \KwIn{Graph $\mathcal{G}(\mathcal{V},\mathcal{E})$; A mini-batch $\mathcal{V}_B$ of nodes; depth L}
\KwIn{set of adjacency matrices $\mathbb{A}$; feature matrix $X$; training set $\mathcal{V}_{train}$ and $\mathcal{Y}_{train}$; Graph Transformer Networks $f_{\theta}$; number of layers $L$, }
% number of sampled neighbors at each layer $\{n^{(l)}\}_{l=1}^{l=L}$; $\epsilon$ }
\KwOut{set of adjacency matrices $\mathbb{A}'$}
% Estimate adjacency matrix $A^{(l)}$ from each GT layer using Eq. 4\;
Combine a set of input adjacency matrices into one adjacency matrix $A \gets \cup_{t=1}^{T_e}{A_t}$ \;
\While{}{
    Sample a mini-batch of m target nodes $\{v_i\}_{i=1}^{m}$ from the training set $\mathcal{V}_{train}$\ with corresponding targets $Y_{B}$ \;
    $\mathcal{V}_{B} \gets \{v_i\}_{i=1}^{m}$\;
    $\mathcal{E}_{B} \gets \phi$ \;
    \For{$l=1,2, \ldots ,L+1$}{
    % $r^{(l)}\gets \phi$\;
    Sample nodes $\mathcal{V}^{(l)}$ and edges $\mathcal{E}^{(l)}$ based on $\mathcal{V}_{B}$ by using graph sampling algorithm \cite{hamilton2017inductive,hu2020hgt} \;
    $\mathcal{V}_{B} \gets \mathcal{V}_{B} \cup \mathcal{V}^{(l)}$\;
    $\mathcal{E}_{B} \gets \mathcal{E}_{B} \cup \mathcal{E}^{(l)}$ \;
    % \For{each node $u \in r^{(l-1)}$}{
    %     Sample $n^{(l)}$ nodes $\{v_i\}_{i=1}^{n^{(l)}}$ from $\mathcal{N}_{u}$ using probability matrix $P^{(l)}$\;
    %     \For{$v \in \{v_i\}_{i=1}^{n^{(l)}}$}{
    %         $r^{(l)} \gets r^{(l)} \cup \{v\}$\;
    %         $\mathcal{E}_{B} \gets \mathcal{E}_{B} \cup \{e_{uv}\}$\;
    %     }
    % }
    }
    Reconstruct an adjacency matrix $A'$ based on sampled nodes $\mathcal{V}_{B}$ and edges $\mathcal{E}_{B}$\;
    Divide the adjacency matrix $A'$ into a set of adjacency matrices corresponding to each edge type $\{A'_t\}_{t=1}^{T_e}$ \;
    Compute prediction $Y'$ from $f_{\theta}$, $X$ and $A'$ by using Eq. (8) and the node classifier\;
    Calculate cross-entropy loss from $Y'$ and $Y_{B}$ \;
    Update weights of $f_{\theta}$ \;
}
% Calculate probability matrix $P^{(l)} \gets \epsilon D^{-1}A + (1-\epsilon) A^{(l)}$ for $l=0,...,L$\;

% Reconstruct a set of sampled adjacency matrix $\mathbb{A}'$ based on sampled edges $\mathcal{E}_{B}$\;
    % $\mathbb{A} \gets \textbf{0}$\;
    % \For{each edge $e_{ij} \in \mathcal{E}_{B}$}{
    %     $t \gets f_e(e_{ij})$\;
    %     $\mathbb{A}_{t}[i,j] = 1$\;
    % }
\caption{Mini-batch training algorithm for Graph Transformer Networks}
\end{algorithm}

\subsection{Proofs of the Proposition 1}
\label{app:Proofs of the Proposition1}
\textbf{Proposition 1. }Given two normalized adjacency matrices $A$, $B \in \Rb^{N \times N}$, the followings are equivalent:
\begin{enumerate}[label=(\roman*)]
    \item $\left( {D}_{A}^{-1}{A}\right) \left({D}_{B}^{-1}{B}\right) = \left({D}_{AB}^{-1}{AB} \right)$
    \item ${D}_{AB}^{-1} = I$
    \item ${D}_{A+I}^{-1}=({D}_{A}+I)^{-1}=\cfrac{1}{2}~I$
\end{enumerate}

% \begin{proof}
Since the adjacency matrices A, B are normalized, i.e., $\sum_{j} A_{ij}=1$, the degree matrices $D_{A}, D_{B}$ are equal to the identity matrix ($I$) and the inverse degree matrices $D_{A}^{-1}, D_{B}^{-1}$ are also equal to the identity matrix ($I$) respectively.
Thus, (\romannumeral 1) in Proposition 1 can be re-written as $AB = {D}_{AB}^{-1}{AB}$~ and to satisfy the (\romannumeral 1), we need to prove that $D_{AB}$ is an identity matrix.
Since $D_{AB}$ is the degree matrix of the multiplication of two matrices $A$ and $B$, each $i$-th diagonal element of $D_{AB}$ can be represented as $D_{AB}[i,i] = \sum_{j}~(AB)_{ij} = \sum_{j}\sum_{k}~A_{ik}B_{kj}$.
Then, we can derive that $D_{AB}$ is equal to the identity matrix, i.e., $D_{AB}[i,i] = 1$ as follows:
\begin{align}
\centering
    D_{AB}[i,i] &=\sum_{j}~(AB)_{ij} \nonumber\\
                &= \sum_{j}\sum_{k}~A_{ik}B_{kj} \nonumber\\
                &= \sum_{k}\sum_{j}~A_{ik}B_{kj} &&\because\text{commutativity of sum} \nonumber\\
                &= \sum_{k}~A_{ik}\sum_{j}~B_{kj} \nonumber \\
                &=  \sum_{k}\left( A_{ik}  \sum_{j}~B_{kj} \right)  \nonumber \\
                &=  \sum_{k}~A_{ik} \nonumber  &&\because\text{$\sum_{j}~B_{kj} = 1$} \\
                &= 1 \nonumber  &&\because\text{$\sum_{k}~A_{ik} = 1$}
\end{align}
Therefore, degree matrix $D_{AB}$ is equal to an identity matrix $I$, which satisfies (\romannumeral 1), (\romannumeral 2) and (\romannumeral 3) in Proposition 1.

\begin{figure*}[t]
 \centering
 \includegraphics[width=350px]{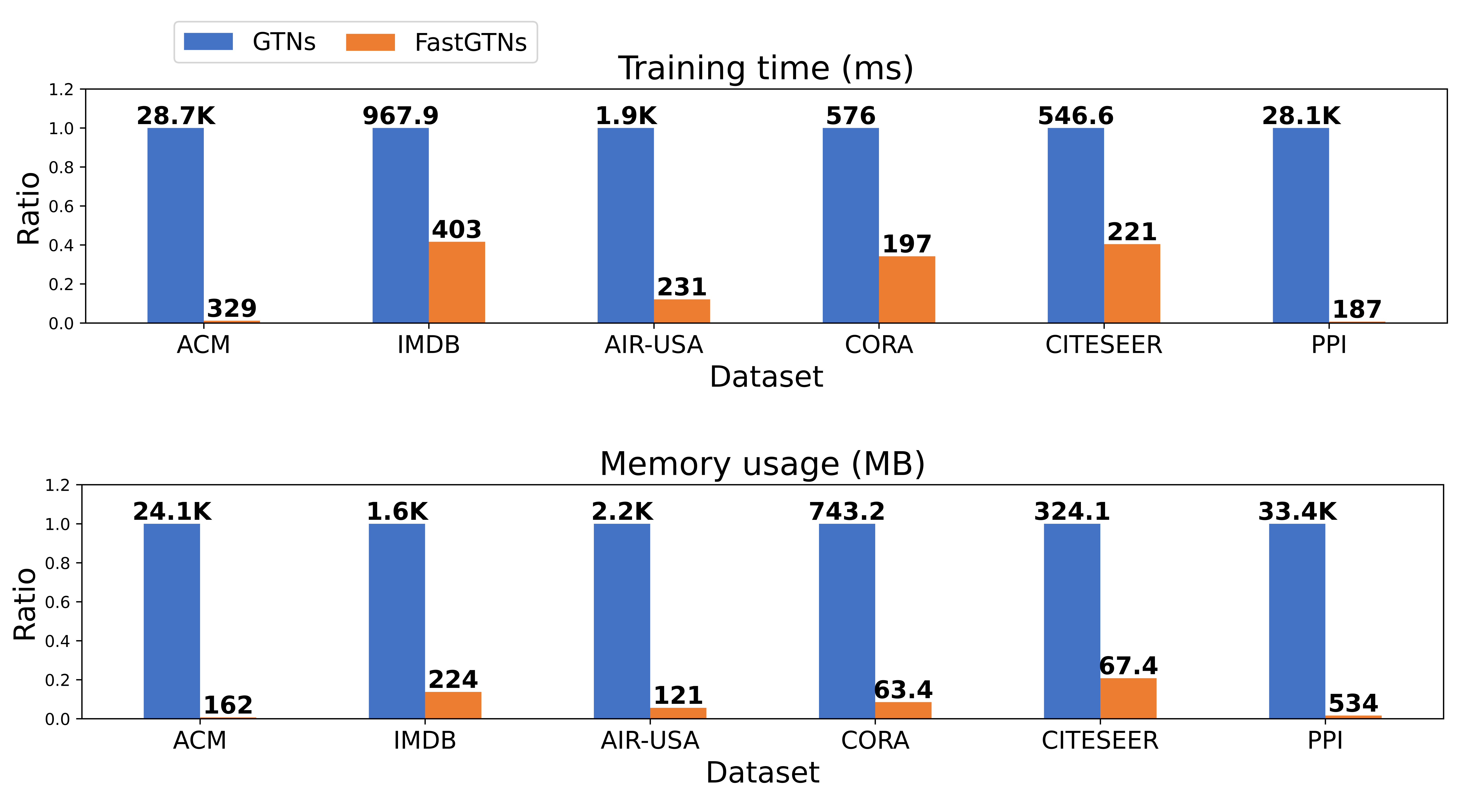}
 \caption{Comparisons of training time (up) and memory usage (down) between GTNs (blue) and FastGTNs (orange) on both homogeneous and heterogeneous graph datasets (x-axis). For fair comparison, FastGTNs were measured without the non-local operations. FastGTNs significantly speed up the graph transformations 150$\times$ (PPI) and reduce the memory usage 60$\times$ (PPI).}
 \label{fig:supp_training_efficiency}
\end{figure*}

\subsection{Relation to RGCN}
\label{app:relation to other GNNs}
If input graphs are normalized, RGCN \cite{schlichtkrull2018rgcn} can be subsumed by our FastGTNs with minor modifications.
The RGCN \cite{schlichtkrull2018rgcn} extends the GCN to heterogeneous graphs by utilizing relation-specific parameters. Specifically, the output representations from the $l$-th RGCN are as 
\begin{equation}
    Z^{(l+1)} = \sigma{\left(\sum_{t=1}^{|\mathcal{T}_e|}{D^{-1}_{t}A_t Z^{(l)} W_t^{(l)}}\right)},
\label{eq:RGCN_appendix}
\end{equation}
where $A_t$ denotes an adjacency matrix of $t$-th edge type $W_t^{(l)}$ denotes the relation specific parameters of the model. The RGCN also address overparameterization by proposing basis decomposition of $W_t^{(l)}$ as $W_t^{(l)}=\sum_{b=1}^B{a_{tb}^{(l)}V_b^{(l)}}$, consequently the equation of the RGCN is re-written as 
\begin{equation}
    Z^{(l+1)} = \sigma{\left(\sum_{t=1}^{|\mathcal{T}_e|}{D^{-1}_{t}A_t Z^{(l)} \left(\sum_{i=1}^B{a_{ti}^{(l)}V_i^{(l)}}\right)}\right)},
\label{eq:RGCN2_appendix}
\end{equation}
Then to compare with our FastGTNs, we derive the equation similar to our FastGTNs as follows:
\begin{align}
Z^{(l+1)} 
          &= \sigma{\left(\sum_{t=1}^{|\mathcal{T}|}{D^{-1}_{t}A_t Z^{(l)} \left(\sum_{i=1}^B{a_{ti}^{(l)}V_i^{(l)}}\right)}\right)} \\
          &= \sigma{\left(\sum_{t=1}^{|\mathcal{T}|}{A_t Z^{(l)}  \left(\sum_{i=1}^B{a_{ti}^{(l)}V_b^{(l)}}\right)}\right)} \\ % &&\because D^{-1}_t=I \\
          &= \sigma{\left(\sum_{t=1}^{|\mathcal{T}|}\sum_{i=1}^B{A_t Z^{(l)} {a_{tb}^{(l)}V_b^{(l)}}}\right)} \\
          &= \sigma{\left(\sum_{i=1}^B \sum_{t=1}^{|\mathcal{T}|} {A_t Z^{(l)} {a_{ti}^{(l)}V_i^{(l)}}}\right)} \\ % &&\because\text{commutativity of sum} \\
          &= \sigma{\left(\sum_{i=1}^B \sum_{t=1}^{|\mathcal{T}|} a_{ti}^{(l)}{A_t Z^{(l)} {V_i^{(l)}}}\right)} \\
          &= \sigma{\left(\sum_{b=1}^B {\left(\sum_{t=1}^{|\mathcal{T}|} {a_{ti}^{(l)} A_t} \right) Z^{(l)} {V_i^{(l)}}}\right)} \\
          &= \sigma{\left(\sum_{i=1}^B a_{i}^{(l)} \cdot {\mathbb{A} Z^{(l)} {V_i^{(l)}}}\right)} \label{eq:RGCN_reformulation}
\end{align}

\subsection{Training Efficiency}
In this section, we compared our FastGTNs with GTNs in terms of the training efficiency. As shown in Figure~\ref{fig:supp_training_efficiency}, we measured the training time and the memory consumption of the two methods. Figure~\ref{fig:supp_training_efficiency} shows that FastGTNs are significantly more efficient than GTNs in terms of both the training time and memory. In particular, the comparison on the large-scale dataset, \textsc{PPI}, shows our FastGTNs are 150$\times$ faster and 60$\times$ more efficient than the GTNs.

\begin{figure}[h]
 \centering
 \includegraphics[width=150px]{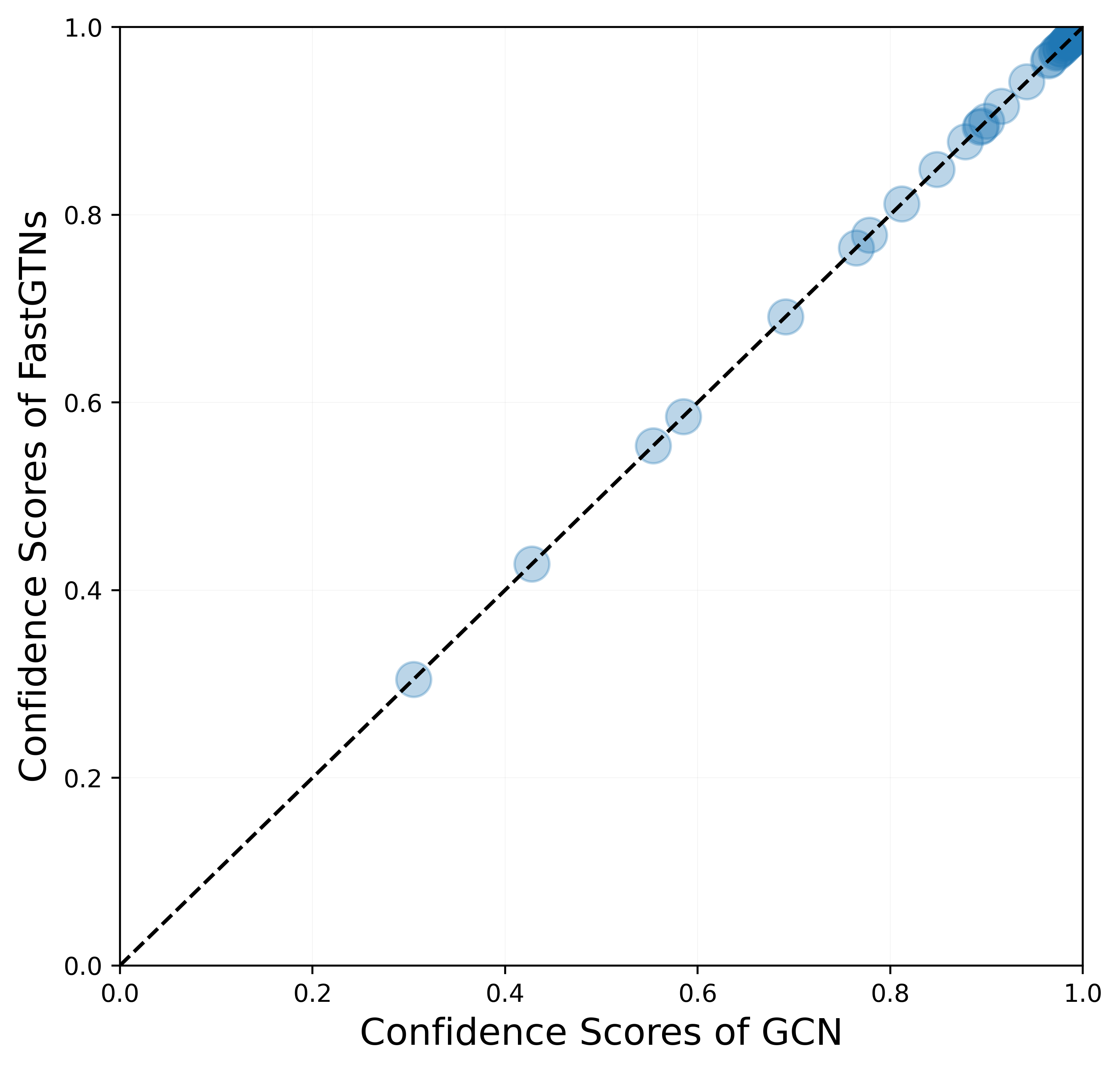}
 \caption{Correlations of confidence scores between GCNs (x-axis) and FastGTNs (y-axis) on 50 randomly drawn data points from a test set of \textsc{CORA} dataset.}
 \label{fig:gcn_confidence scores}
\end{figure}
\begin{figure}[h]
 \centering
 \includegraphics[width=150px]{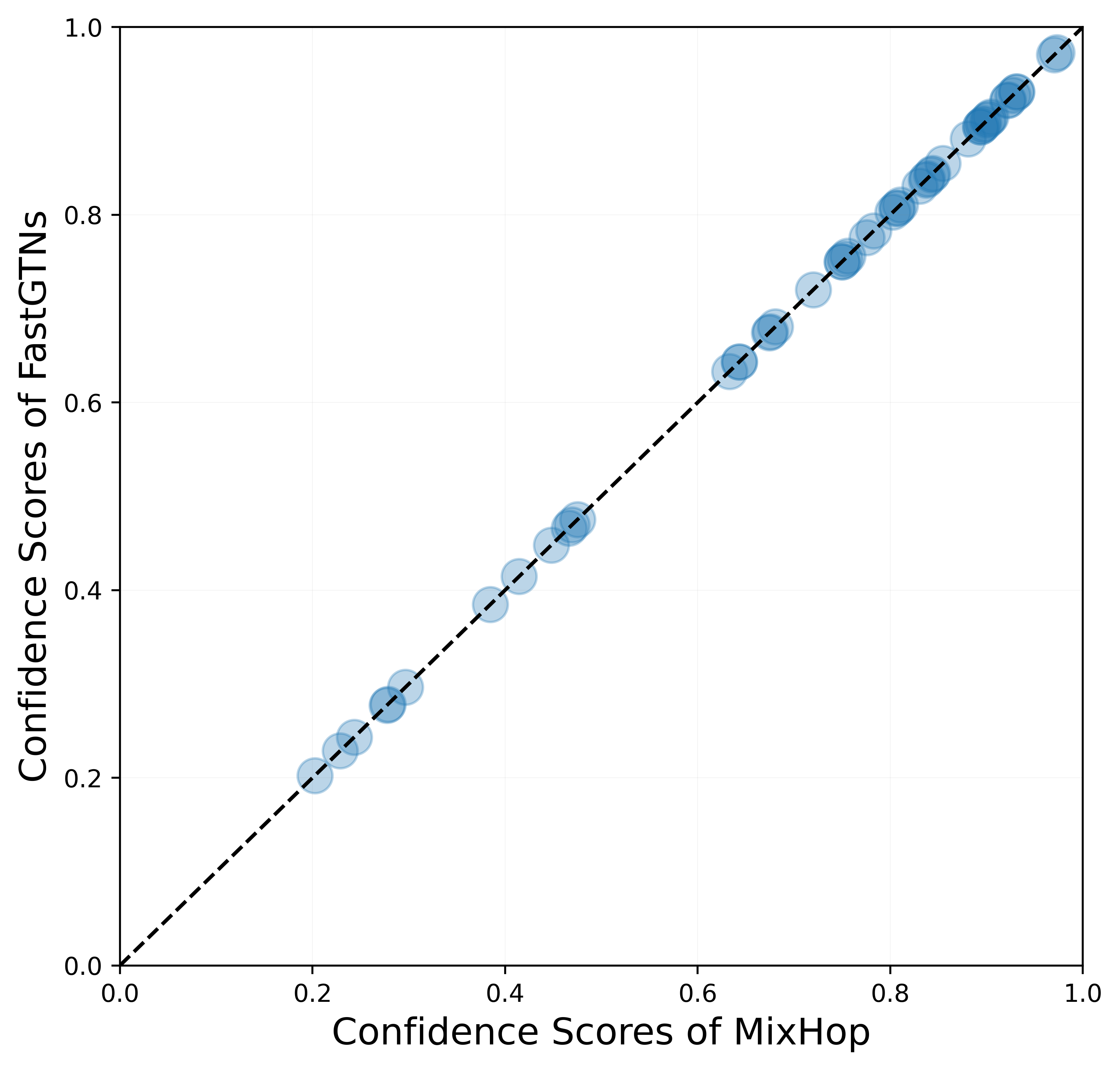}
 \caption{Correlations of confidence scores between MixHop (x-axis) and FastGTNs (y-axis) on 50 randomly drawn data points from a test set of \textsc{CORA} dataset.}
 \label{fig:mixhop_confidence scores}
\end{figure}

\subsection{Generalization of GCN and MixHop}
\label{fig:supp_subsumption}
As discussed in Section 3.6, our FastGTNs subsume two popular graph neural networks, GCN and MixHop. 
For the GCN, if the number of FastGT layers in FastGTNs is one i.e., $K=1$, the number of channels is one, i.e., $C=1$, $\gamma$ equals $\frac{1}{2}$ and the first FastGT layer only selects the adjacency matrix, i.e., $\alphab^{(1)}\cdot \mathbb{A} = 1 \cdot A+0 \cdot I$, the GCN can be a special case of FastGTNs.
For the MixHop, if $\gamma$ equals $0$, the number of channels equals the size of $P$ i.e., $C=|P|$, number of FastGT layers in each channel equals $j$ and all FastGT layers choose the adjacency matrix and the identity matrix in the same ratio, i.e., $\alphab^{(k)}\cdot \mathbb{A} = \frac{1}{2}A+ \frac{1}{2}I$, the output is same as the output of the Mixhop, i.e., the Mixhop can be a special case of FastGTNs.
To show that special cases of FastGTNs can be identically performed by the GCN and the MixHop, we first train a GCN and a MixHop on the \textsc{CORA} dataset and copy the model parameters of GCN($\{W^{(l)}\}_{l=1}^L$) and MixHop($\{W_{j}^{(l)}\}_{l=1}^L$) to the corresponding model parameters in our FastGTN. 
Figure \ref{fig:gcn_confidence scores} and Figure \ref{fig:gcn_confidence scores} prove that the predictions (confidence scores) by the FastGTN and other graph neural networks, GCN and MixHop, are identical. All 50 randomly drawn data points from a test set are on the Identity line (i.e., $y=x$).

\subsection{Ablation Study for Non-local Operations}
\label{app:ablation_study}
We evaluate the effectiveness of non-local operations in FastGTNs. 
Table \ref{tab:non-local} shows the performance gap of FastGTNs with/without non-local operations. 
Also, the attention scores on the non-local adjacency matrix at each layer of FastGTNs 
are reported to show whether the non-local operations are adaptively applied.
First, non-local operations improve the performance in most datasets (7 out of 9) by $0.39\sim4.4$ in terms of the micro-F1 score. 
In addition, Table \ref{tab:non-local} shows that if the non-local operations are useful then FastGTNs have relatively higher attention scores on non-local operations. 
In contrast, in \textsc{Blogcatalog}, \textsc{Flickr} datasets, FastGTNs where non-local operations are not useful, the FastGTNs (with non-local operations) properly reduced the attention scores on non-local operations to $0.004 \sim 0.012$. 
The results show that FastGTNs can adaptively exploit non-local operations depending on datasets.

\begin{table}[h]
\centering
% \begin{adjustbox}{width=0.48\textwidth}
\begin{tabular}{ l c c c c c }
\toprule
\multicolumn{1}{ c }{\multirow{2}{*}{Dataset}} & {\multirow{2}{*}{\shortstack{w/o \\ non-local}}} & \multirow{2}{*}{\begin{tabular}[c]{@{}c@{}}\shortstack{w/ \\ non-local}\end{tabular}} & \multicolumn{2}{c }{\begin{tabular}[c]{@{}c@{}} Attention scores \\ on non-local op. \end{tabular}} \\ \cline{4-5} 
\multicolumn{1}{ c }{} &  &  & L1 & L2 \\ \midrule % & L3 \\ \midrule
\textsc{Blogc} & \textbf{88.96} & 87.97 ~($\downarrow$ 0.99) & 0.012 & 0.005 \\ %& - \\
\textsc{Flickr} & \textbf{75.01} & 73.64 ~($\downarrow$ 1.37) & 0.004 & 0.005 \\ %& - \\  
\textsc{Air-USA} & 56.60 & \textbf{57.73} ~($\uparrow$ 1.13) & 0.153 & 0.158 \\ %& 0.142 \\ 
\textsc{Citeseer} & 68.32 & \textbf{69.14} ~($\uparrow$ 0.82) & 0.099 & 0.095 \\ %& 0.089 \\ 
\textsc{Cora} & 79.17 & \textbf{80.29} ~($\uparrow$ 1.12) & 0.146 & 0.152 \\ %& 0.156 \\
\textsc{PPI} & 40.95 & \textbf{42.40} ~($\uparrow$ 1.45) & 0.243 & 0.223 \\ \midrule %& 0.248 \\ \midrule
DBLP & 94.46 & \textbf{94.85} ~($\uparrow$ 0.39) & 0.048 & 0.049 \\ %& 0.047 \\
ACM & 91.79 & \textbf{92.51} ~($\uparrow$ 0.72) & 0.026 & 0.026 \\ %& 0.025 \\ 
IMDB & 60.23 & \textbf{64.63} ~($\uparrow$ 4.4) & 0.066 & 0.069 \\ \bottomrule %& 0.069 \\ \bottomrule
\end{tabular}
% \end{adjustbox}
\vspace{1mm}
\
\caption{Ablation study for non-local operations on both homogeneous and heterogeneous graph datasets. Non-local operations improve the performance of FastGTNs on all datasets except for on \textsc{Blogcatalog} and \textsc{Flickr} datasets. The attention scores on non-local operations show that our FastGTNs adaptively leverage non-local operations by adjusting the attention scores on non-local adjacency matrices.}
\label{tab:non-local}
\end{table}

\subsection{Interpretation of GTNs in Homogeneous Graphs}
\label{appendix:attention_scores_homogeneous}
\begin{figure}[h]
    \centering
    \subfloat[\textsc{Air-usa}]{\includegraphics[trim={0, 5, 0, 5}, width=7cm, height=4cm]{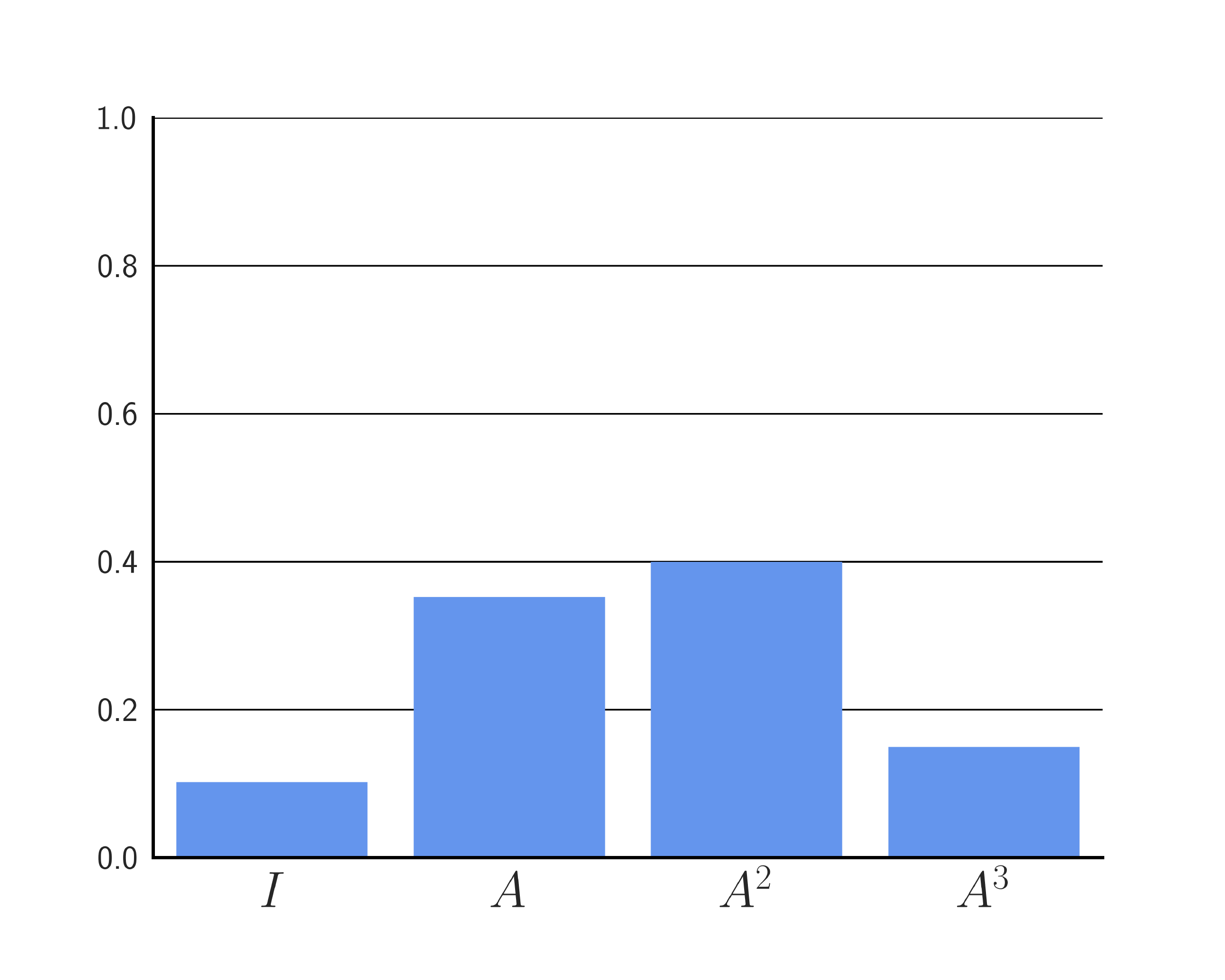}} \\
    % \hfill
    \subfloat[\textsc{BlogCatalog}]{\includegraphics[trim={0, 5, 0, 5}, width=7cm, height=4cm]{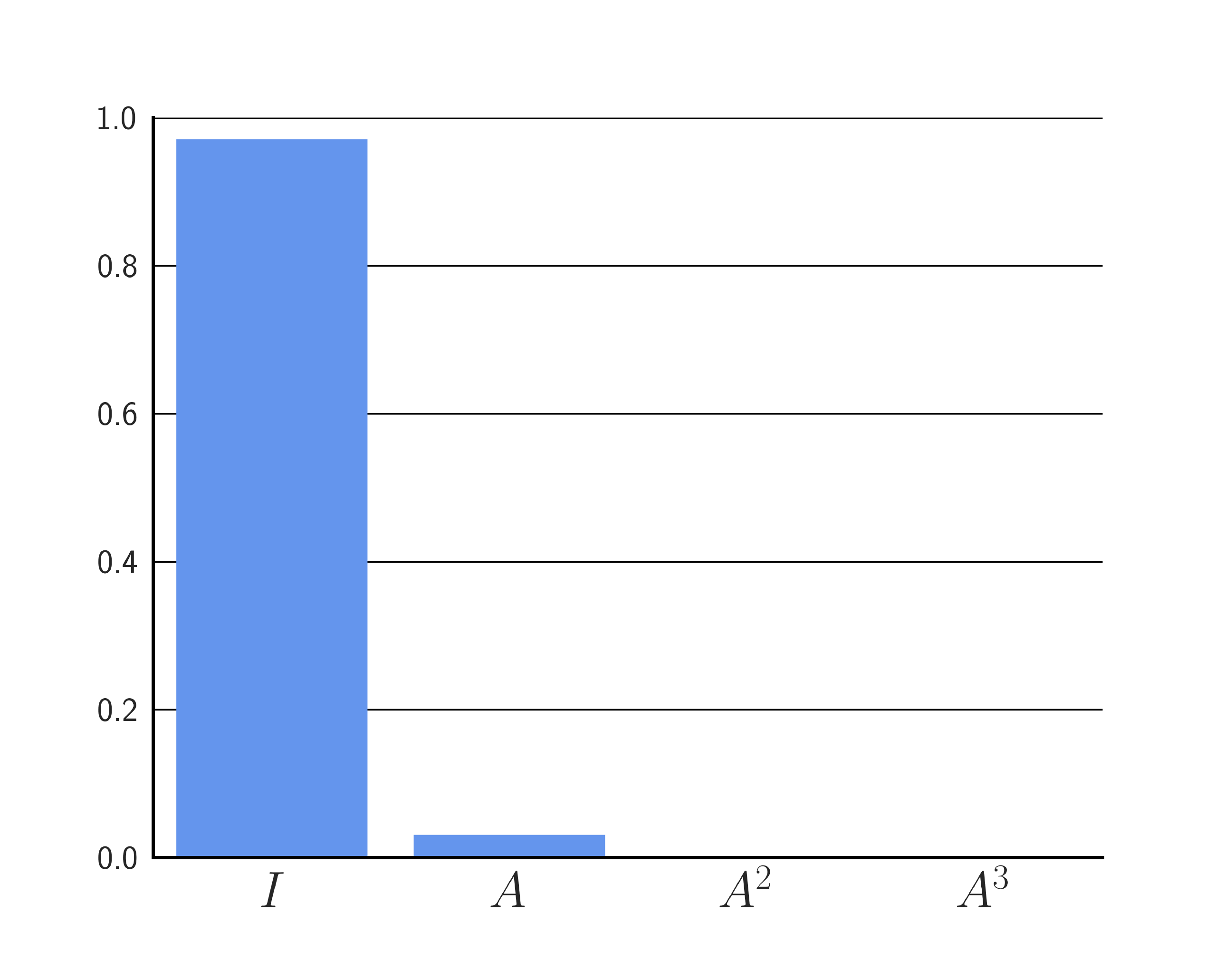}}
    % \begin{subfig}{\columnwidth}
    %     \includegraphics[trim={0, 0, 0, 0}, width=\columnwidth]{figures/gtn_plot2.pdf}
    %     \subcaption{Thing}
    % \end{subfig}
    \caption{
    We visualized the ratios corresponding to each hop in the output matrix ${A}^{(l)}$ from GTNs based on the attention scores of each GT Layer in \textsc{Air-usa} (left) and \textsc{BlogCatalog} (right) dataset. 
    In (a), we use two GT layers in GTNs, thus output matrix from GTNs can involve up to three hop adjacency matrix. In (b), we use one GT layer, thus output matrix from GTNs can involve up to two hop adjacency matrix.
    }
    \label{fig:homogeneous_attention_score}
\end{figure}

Fig.~\ref{fig:homogeneous_attention_score} shows ratios of each power of adjacency matrix in the output matrix ${A}^{(K)}$ from GTNs, respectively in the \textsc{Air-usa} (left) and \textsc{BlogCatalog} (right) datasets.
% Based on the attention scores of each GT Layer, we calculate the ratio of each hop in the final
Based on the Eq. (30), we calculate the ratio of each hop in the final adjacency matrix after GTNs. Specifically, if the number of GT layers equals two, then the ratio of the identity matrix $I$ is obtained by $(\alpha_{I}^{(0)}*\alpha_{I}^{(1)}*\alpha_{I}^{(2)})$ and the ratio of the adjacency matrix $A$ is obtained by $(\alpha_{A}^{(0)}*\alpha_{I}^{(1)} *\alpha_{I}^{(2)} + \alpha_{I}^{(0)}*\alpha_{A}^{(1)}*\alpha_{I}^{(2)} + \alpha_{I}^{(0)}* \alpha_{I}^{(1)}*\alpha_{A}^{(2)})$. As we discussed in Section 4.2, since graph structures in the \textsc{BlogCatalog} dataset are noisy enough for a simple MLP to outperform other GNN baselines, GTNs learn to assign higher attention scores to the identity matrix, which effectively minimizes the range of neighborhoods. In contrast, in the \textsc{Air-usa} dataset, each GT layer assigns relatively higher attention scores to the adjacency matrix, which expands the range of neighborhoods. GTNs can adaptively learn effective range of neighborhoods depending on the dataset.

\end{document}